\documentclass[letterpaper, 10 pt, journal, twoside]{IEEEtran}

\usepackage{amsmath,amsfonts}
\usepackage{algorithmic}
\usepackage{array}
\usepackage[caption=false,font=normalsize,labelfont=sf,textfont=sf]{subfig}
\usepackage{textcomp}
\usepackage{stfloats}
\usepackage{url}
\usepackage{verbatim}
\usepackage{graphicx}
\def\BibTeX{{\rm B\kern-.05em{\sc i\kern-.025em b}\kern-.08em
    T\kern-.1667em\lower.7ex\hbox{E}\kern-.125emX}}
\usepackage{balance}

\usepackage{comment}
\usepackage{xcolor}
\usepackage{multirow}
\usepackage{times}
\usepackage{epsfig}
\usepackage{amsmath}
\usepackage{amssymb}
\usepackage{bm}
\usepackage{siunitx}
\usepackage{graphics} 
\usepackage{graphicx}
\usepackage{float}
\usepackage{mathrsfs}
\usepackage{booktabs}
\usepackage{cite}
\usepackage{helvet}
\usepackage{fontawesome5}

\definecolor{lightviolet}{rgb}{0.722, 0.549, 0.867}
\definecolor{lightred}{rgb}{1.0, 0.388, 0.388}
\usepackage[pagebackref=false,breaklinks=true,colorlinks=true,bookmarks=false,linkcolor=lightred,citecolor=cyan,urlcolor=lightviolet]{hyperref}

\newcommand{\red}[1]{\textcolor{red}{#1}}

\newcommand{\black}[1]{\textcolor{black}{#1}}

\newcommand{\FMCWLIO}{%
\sffamily\bfseries
\textcolor[rgb]{0.945,0.149,0.114}{F}%
\textcolor[rgb]{1.000,0.573,0.000}{M}%
\textcolor[rgb]{0.992,0.859,0.165}{C}%
\textcolor[rgb]{0.302,0.894,0.169}{W}%
\textcolor[rgb]{0.059,0.965,0.573}{-}%
\textcolor[rgb]{0.043,0.847,0.976}{L}%
\textcolor[rgb]{0.173,0.133,0.984}{I}%
\textcolor[rgb]{0.765,0.176,0.953}{O}%
}

\newcommand{\FreeInit}{%
{\sffamily\bfseries\itshape\color[rgb]{0.95,0.35,0.35}Free-Init}%
}

\begin{document}


\title{
{\href{https://doi.org/10.1109/LRA.2024.3396636}{\FMCWLIO}} \\ {\vspace{5pt}}
A Doppler LiDAR-Inertial Odometry
}

\author{Mingle Zhao, Jiahao Wang, Tianxiao Gao, Chengzhong Xu, and Hui Kong {\vspace{-8pt}}
\thanks{The authors are with the University of Macau.}
\thanks{\textbf{Publication}: M. Zhao, J. Wang, T. Gao, C. Xu and H. Kong, "FMCW-LIO: A Doppler LiDAR-Inertial Odometry," in \textit{IEEE Robotics and Automation Letters}, vol. 9, no. 6, pp. 5727-5734, 2024, DOI: {\href{https://doi.org/10.1109/lra.2024.3396636}{10.1109/LRA.2024.3396636}}.}
}


\maketitle



\begin{abstract}
Conventional LiDAR-inertial odometry (LIO) or simultaneous localization and mapping (SLAM) methods heavily rely on geometric features of environments, as LiDARs primarily provide range measurements instead of motion measurements. From now on, however, the situation changes thanks to the novel Frequency Modulated Continuous Wave (FMCW) Doppler LiDARs. FMCW Doppler LiDARs not only offer the point range with high resolution but also capture the instant point Doppler velocity through the Doppler effect. In the letter, we propose {\href{https://doi.org/10.1109/LRA.2024.3396636}{\FMCWLIO}}, a novel and robust LIO, leveraging intrinsic Doppler measurements from FMCW Doppler LiDARs. To correctly exploit Doppler velocities, a motion compensation method is designed, and a Doppler-aided observation model is applied for on-manifold state estimation. Then, dynamic points can be effectively removed by the Doppler criteria, deriving more consistent geometric observations. FMCW-LIO eventually achieves accurate state estimation and static mapping, even in structure-degenerated environments. Extensive experiments in diverse scenes are performed and FMCW-LIO outperforms other algorithms on both accuracy and robustness. 
\end{abstract}




\phantomsection
\section*{Resources} \label{sec:resources}
\vspace{5pt}

    \begin{center}
    \begin{tabular}{@{}l@{\hspace{8pt}}c@{\hspace{8pt}}l@{}}
        IEEE Xplore Link & : & {\href{https://ieeexplore.ieee.org/document/10518074}{\FMCWLIO}} \\
        arXiv Paper Link & : & {\href{https://arxiv.org/abs/2609.29374}{{\red{\faFilePdf}}{\enspace{FMCW-LIO}}}} \\
        Code \& Dataset & : & {\href{https://github.com/IMRL/FMCW-LIO}{{\black{\faGithub}}{\enspace{FMCW-LIO}}}} \\
        Experiment Video & : & {\href{https://youtu.be/2yuZYw91AP8}{{\red{\faYoutube}}{\enspace{FMCW-LIO}}}}
    \end{tabular}
    \end{center}


\section{Introduction} \label{sec:intro}

\IEEEPARstart{L}{aser} Light Detection And Ranging (LiDAR) sensors are indispensable and play a crucial role in robotics and automation. In recent years, LiDAR sensors have undergone dramatic innovations with more advanced features. Among them, a remarkable milestone is the technology of Frequency Modulated Continuous Wave (FMCW) Doppler LiDAR that can capture both the range and radial Doppler velocity of points based on the modulation and demodulation of laser waves in frequency domain \cite{anderson2017ladar} \cite{DICP}. Additionally, the FMCW Doppler LiDAR shows promise of miniaturization, thus holding significant potential in robotic applications. 

The new dimension of Doppler velocity in the FMCW Doppler LiDAR implicitly conveys the motion information of the LiDAR, which opens up new pathways for robotic perception and state estimation. Consequently, numerous unexplored topics arise in the integration of conventional methods, previously dominated by geometric feature-based methods, with the intrinsic Doppler velocity. Besides, FMCW Doppler LiDARs can output precise range sensing of massive points with significantly high temporal-spatial resolution. On the other hand, FMCW Doppler LiDARs possess the inherent capability to measure the Doppler velocity. More importantly, Doppler measurements can provide a direct observation of the LiDAR motion state (LiDAR velocity), which has a correspondence-free nature. The correspondence-free observation is irrelevant to the locally geometric features (e.g., line or plane features) of the surrounding environment. As a result, FMCW Doppler LiDARs, representing the next generation of laser sensing technology, hold the potential to address various challenging issues in the conventional LiDAR field, including the degeneracy and dynamic problem \cite{DICP}.

    \begin{figure}[!t]
       \centering
       \includegraphics[width=0.47\textwidth]{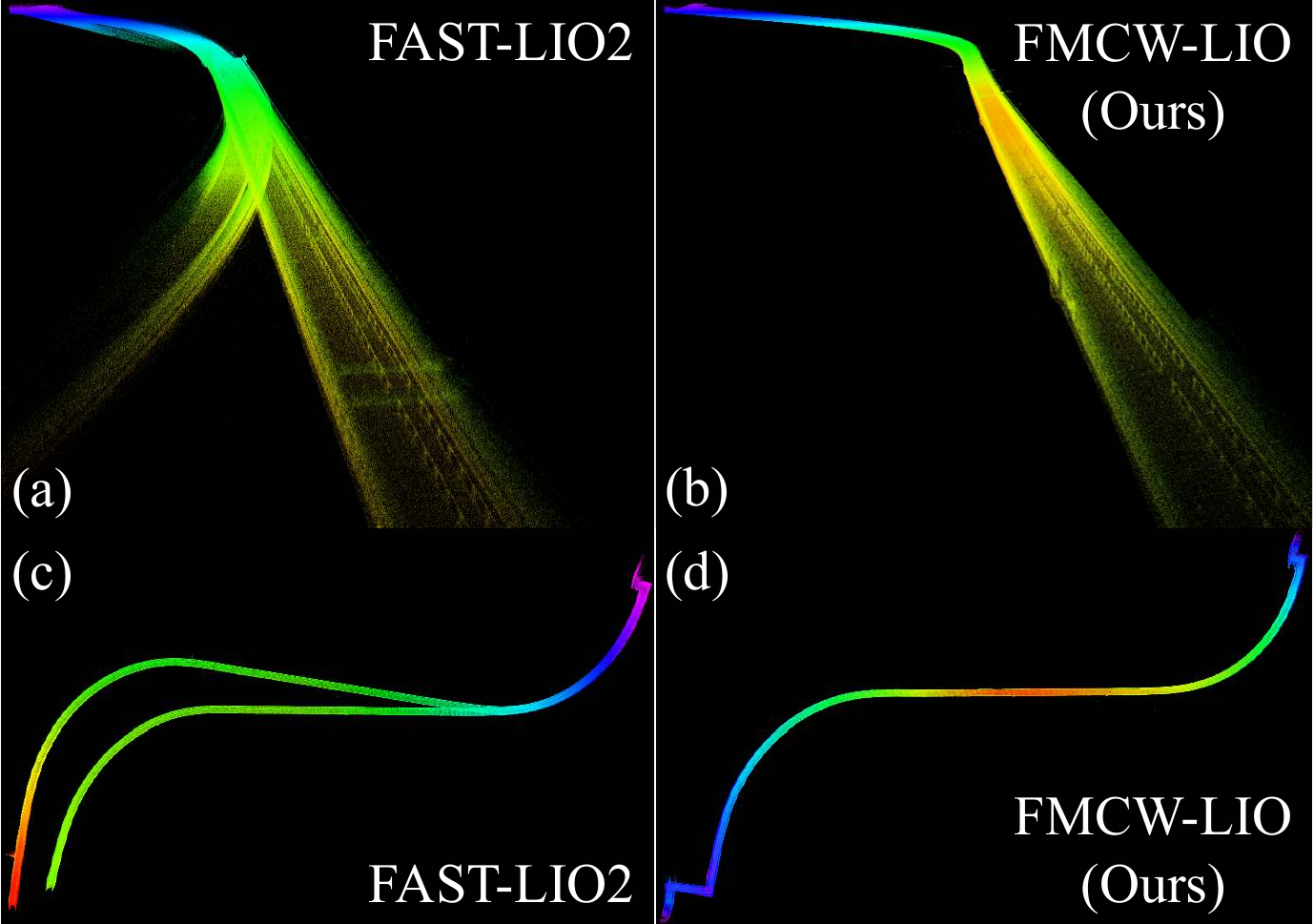}
       \caption{Online mapping comparison of the whole S-shaped tunnel. (a) and (c) show the drifted and inconsistent maps of FAST-LIO2. (b) and (d) show the neat and consistent maps of FMCW-LIO.}
       \label{fig:system_demo}
    \end{figure}

In this work, harnessing the advantages of FMCW Doppler LiDAR and Inertial Measurement Unit (IMU), we propose a Doppler LiDAR-inertial odometry and static mapping framework. The contributions can be summarized as follows: 
\begin{enumerate}
    \item A robust Doppler-aided LIO, {\href{https://doi.org/10.1109/LRA.2024.3396636}{\FMCWLIO}}, is proposed for FMCW Doppler LiDAR systems. To our knowledge, FMCW-LIO is the first framework for FMCW Doppler LiDAR-inertial state estimation and static mapping. 
    
    \item To precisely utilize Doppler velocities in an accumulated LiDAR scan, a novel motion compensation method for Doppler measurements is designed by fusing with IMU. Notably, beyond motion compensation, the proposed Doppler relationship essentially reveals the interrelation between Doppler velocities of a static point with respect to different LiDAR frames.

    \item A Doppler-aided observation, integrated in the on-manifold Iterated Error-State Extended Kalman Filter (IESEKF), and an efficient dynamic point removal method are exploited to obtain consistent geometric observations and static maps. 

    \item Comprehensive experiments on multiple mobile platforms equipped with an FMCW Doppler LiDAR are conducted to validate and demonstrate the effectiveness, accuracy, and robustness of the proposed framework. 
    
    \item The source code of FMCW-LIO, the data sequences, the integrated framework of FMCW-LIO with {\href{https://doi.org/10.1109/LRA.2024.3490395}{\FreeInit}}, and the experiment video are made publicly available (see the \nameref{sec:resources} section for details).
\end{enumerate}


\section{Related Work} \label{sec:relatedwork}

In this section, we review the most related works focusing on LiDAR-inertial odometry, FMCW Doppler LiDAR-based algorithms and datasets, and recent works on automotive radar-inertial odometry. 
\subsection{LiDAR-Inertial Odometry}

To handle dynamic motions of robots, the tightly-coupled LiDAR-inertial fusion has attracted significant research attention. LIO-SAM \cite{LIO-SAM} employs a factor-graph framework, and it can also fuse Global Navigation Satellite System (GNSS) and loop closure factors. LIO-mapping \cite{LIO-Mapping} uses the IMU pre-integration in a constrained refinement on estimation and mapping. LINS \cite{LINS} introduces a framework based on the IESEKF to recursively update the state, but it heavily relies on the ground assumption. In \cite{FAST-LIO}, authors apply the IESEKF and a novel Kalman gain formulation for efficient computing. FAST-LIO2 \cite{FAST-LIO2} further enhances the performance by using a direct scan-to-map observation model and the incremental k-dimensional tree (\textit{ikd-Tree}) for the online map. Faster-LIO \cite{Faster-LIO} uses an incremental voxel map (iVox) to achieve higher search speed. Point-LIO \cite{Point-LIO} uses a point-wise framework and a novel IMU modeling method that allows a high-frequency odometry output. A hierarchical geometrical observer and a higher-order motion correction are applied in \cite{DLIO} which can efficiently improve the estimation accuracy and map quality. 
\subsection{FMCW Doppler LiDAR-based Algorithms and Datasets} \label{subsec:fmcw_related_work}

Recently, DICP \cite{DICP} introduces a variant of conventional geometric Iterative Closest Point (ICP) methods by including a Doppler objective function in the registration problem, revealing significant improvements in accuracy even in feature-denied scenes. Guo et al. \cite{guo2022doppler_vel} propose a velocity estimation and clustering framework with Doppler velocities in the CARLA simulation \cite{CARLA}. Gu et al. \cite{gu2022fmcw} present a moving-object tracking method using an FMCW Doppler LiDAR. In \cite{Picking_up_Speed} and \cite{Need_for_Speed}, authors present a continuous-time LiDAR odometry (LO) and a velocity estimator supporting Doppler measurements. In \cite{yoneda2023EOT}, authors use FMCW Doppler LiDARs to simultaneously estimate the kinematic state and the shape of moving objects. More recently, Jung et al. \cite{HeLiPR} release a heterogeneous LiDAR dataset (HeLiPR) for place recognition tasks, including an FMCW Doppler LiDAR. However, HeLiPR is not appropriate for online state estimation because the velocity of points is the post-processed absolute velocity using external sensors, instead of the raw Doppler velocity measurement. 
\subsection{Automotive Radar-Inertial Odometry} 

3D or 4D imaging radars can measure the position and Doppler velocity of points similar to FMCW Doppler LiDARs, but the point density is far less than that of FMCW Doppler LiDARs. In \cite{rio} \cite{michalczyk2022tightly-coupled_rio}, authors propose EKF-based radar-inertial odometry (RIO) systems, with ego-velocity estimates. In \cite{Milli-RIO}, authors use the radar intensity in the RIO. Ng et al. \cite{ng2021continuous} propose a radar-inertial odometry which uses a continuous-time framework. Lately, 4D iRIOM \cite{4D_iRIOM} shows the impressive localization and mapping accuracy of a RIO, integrated with the scan-to-submap matching and loop closure modules.


    \begin{figure}[!t]
       \centering
       \includegraphics[width=0.47\textwidth]{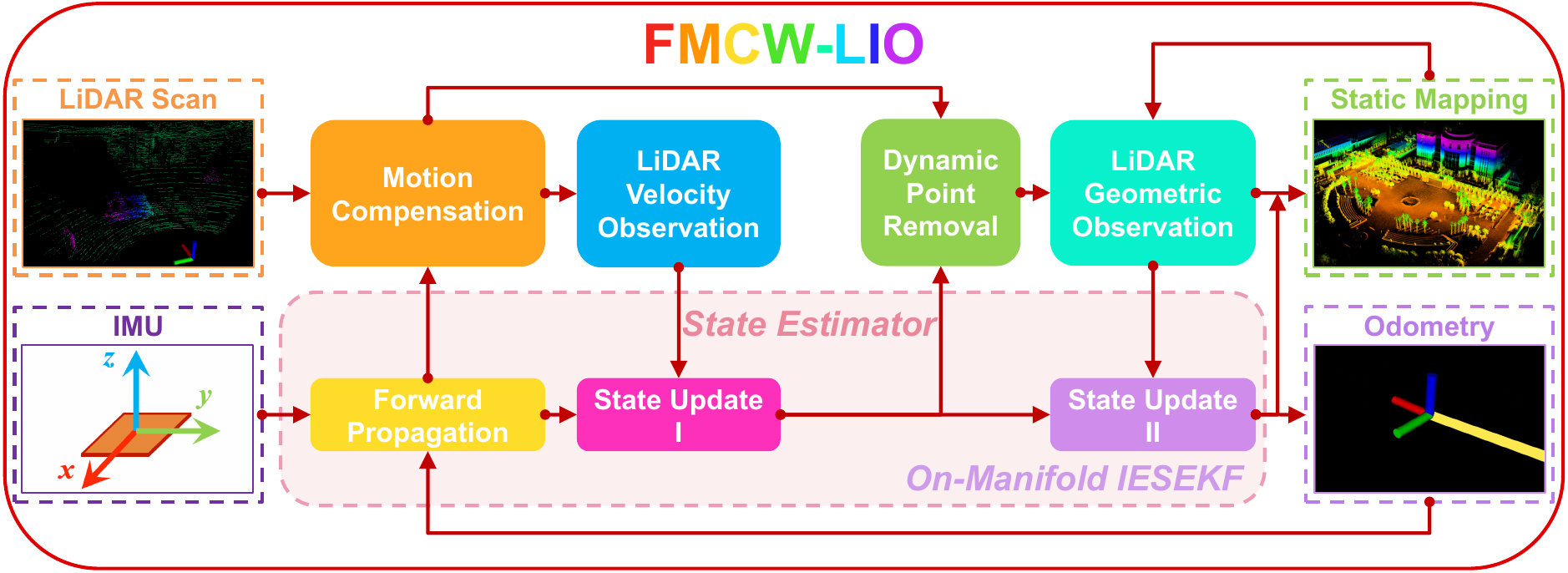}
       \caption{System overview of FMCW-LIO.}
       \label{fig:system_overview}
    \end{figure} 

\section{System Overview} \label{sec:system_overview}

The system overview is illustrated in Fig. \ref{fig:system_overview}. FMCW-LIO takes 6-axis IMU signals as system inputs. LiDAR scans from an FMCW Doppler LiDAR are fed into the motion compensation to correct the distortion of position and Doppler measurements. The undistorted scans are first used in the LiDAR velocity observation model. Adopting the updated velocity, potential dynamic points in the scan can be removed, according to raw Doppler measurements. Then, the remaining static points are sent to the LiDAR geometric observation, resulting in more accurate and consistent associations for the second state update and static mapping. Overall, the Doppler-aided FMCW-LIO can achieve accurate and robust state estimation and output a consistent, neat, and static map. 

For notations, the inertial frame is set as the world frame $w$ and the body frame $b$ is the IMU frame. $^{w}(\cdot)$, $^{b}(\cdot)$, and $^{l}(\cdot)$ respectively represent a 3D vector projected in the world, body, and LiDAR frame. The gravity ${^{w}\mathbf{g}}$ is constant in the world frame under the steady and flat earth assumption. A 3D vector $\mathbf{p}$ from point $A$ to $B$ projected in the frame $C$ is $^{C}\mathbf{p}_{AB}$. Similarly, a 3D linear velocity of the frame $B$ with respect to the frame $A$ expressed in the frame $C$ is $^{C}\mathbf{v}_{AB}$. We use $\mathbf{R} \in SO(3)$ to represent rotations and a rotation matrix rotating a vector in the frame $B$ to $A$ is $\mathbf{R}_{AB}$. $\mathbf{T}_{AB} \in SE(3)$ physically transforms a vector in the frame $B$ to the frame $A$, considering the lever arm. $\mathbf{0}$ is the $3 \times 1$ zero vector, $\mathbf{0}_{n \times m}$ is the $n \times m$ zero matrix, $\mathbf{I}$ is the $3 \times 3$ identity matrix, and $\mathbf{I}_{n}$ is the $n \times n$ identity matrix. $\widetilde{(\cdot)}$, $\widehat{(\cdot)}$, and $\overline{(\cdot)}$ are respectively the measurement, the propagated state, and the updated state. 


\section{Methodology} \label{sec:methodology}

\subsection{System State Description}

The state $\mathbf{x} \in \mathcal{M}$ evolves on the 24-dimensional manifold $\mathcal{M}$, including the rotation $\mathbf{R}_{wb}$ and position $^{w}\mathbf{p}_{wb}$ of body frame with respect to the world frame (the first body frame), velocity $^{w}\mathbf{v}_{wb}$, gyroscope and accelerometer biases $^{b}\mathbf{b}_{g}$ and $^{b}\mathbf{b}_{a}$, and LiDAR-IMU extrinsic parameters $\mathbf{R}_{bl}$ and $^{b}\mathbf{p}_{bl}$: 
    \begin{equation*}
    \setlength{\arraycolsep}{4.2pt}
    \begin{aligned}
        \mathcal{M} &\triangleq SO(3) \times {\mathbb{R}^{15}} \times SO(3) \times {\mathbb{R}^{3}}, \quad \text{dim}(\mathcal{M}) = 24 , \\ 
        \mathbf{x} &\triangleq {\begin{bmatrix} \mathbf{R}_{wb}^{\top} & {^{w}\mathbf{p}_{wb}^{\top}} & {^{w}\mathbf{v}_{wb}^{\top}} & {^{b}\mathbf{b}_{g}^{\top}} & {^{b}\mathbf{b}_{a}^{\top}} & {^{w}\mathbf{g}^{\top}} & \mathbf{R}_{bl}^{\top} & {^{b}\mathbf{p}_{bl}^{\top}} \end{bmatrix}}^{\top} .
    \end{aligned}
    \end{equation*}
To parameterize the error state on the tangent space of the locally homeomorphic manifold, two operators (``boxplus'': $\boxplus$ and ``boxminus'': $\boxminus$) are utilized \cite{FAST-LIO2} \cite{IKFoM-arXiv}. Hence the error-state ${\delta \mathbf{x}} \triangleq \mathbf{x} \boxminus {\mathbf{x}}_{esti} \in \mathbb{R}^{24}$ between the true state $\mathbf{x}$ and the nominal estimate ${\mathbf{x}}_{esti}$ (${\mathbf{x}}_{esti} = {\widehat{\mathbf{x}}}$ or ${\mathbf{x}}_{esti} = {\overline{\mathbf{x}}}$) is: 
    \begin{equation*}
    \setlength{\arraycolsep}{1.8pt}
    \begin{aligned}
        {{\delta} \mathbf{x}} = {\begin{bmatrix} {\delta} {\bm{\theta}_{wb}^{\top}} & {^{w}{\delta} \mathbf{p}_{wb}^{\top}} & {^{w}{\delta} \mathbf{v}_{wb}^{\top}} & {^{b}{\delta} \mathbf{b}_{g}^{\top}} & {^{b}{\delta} \mathbf{b}_{a}^{\top}} & {^{w}{\delta} \mathbf{g}^{\top}} & {\delta} {\bm{\theta}_{bl}^{\top}} & {^{b}{\delta} \mathbf{p}_{bl}^{\top}} \end{bmatrix}}^{\top} .
    \end{aligned}
    \end{equation*}
\subsection{State Propagation} 

\subsubsection{System Input}
The system input is the IMU measurement (angular velocity $^{b}\widetilde{\bm{\omega}}$ and specific force $^{b}\widetilde{\mathbf{a}}$). 
The input is denoted as $\mathbf{u}$, interrupted by zero-mean Gaussian noises $\mathbf{n}_{g}$ and $\mathbf{n}_{a}$. $\mathbf{n}_{bg}$ and $\mathbf{n}_{ba}$ are Gaussian noises of IMU biases, modeled as random walks. The input $\mathbf{u}$ and the noise $\mathbf{w}$ are: 
    \begin{equation*}
    \setlength{\arraycolsep}{3.2pt}
    \begin{aligned}
        \mathbf{u} \triangleq {\begin{bmatrix} {^{b}\widetilde{\bm{\omega}}}^{\top} & {^{b}\widetilde{\mathbf{a}}}^{\top} \end{bmatrix}}^{\top} \in \mathbb{R}^{6}, \  \mathbf{w} \triangleq {\begin{bmatrix} \mathbf{n}_g^{\top} & \mathbf{n}_a^{\top} & \mathbf{n}_{bg}^{\top} & \mathbf{n}_{ba}^{\top} \end{bmatrix}}^{\top} \in \mathbb{R}^{12} .
    \end{aligned}
    \end{equation*}

\subsubsection{Continuous-Time Model} 
Considering the input signal and the rigid connection between LiDAR-IMU frames, the state dynamics model in continuous time can be derived as: 
    \begin{align}
    \label{eq:continuous_model}
        \dot{\mathbf{R}}_{wb} &= \mathbf{R}_{wb} \left[ {^{b}\widetilde{\bm{\omega}}} - {^{b}\mathbf{b}_{g}} - \mathbf{n}_{g} \right]_{\times}, \  {^{w}\dot{\mathbf{p}}_{wb}} = {^{w}\mathbf{v}_{wb}} \notag \\ 
        {^{w}\dot{\mathbf{v}}_{wb}} &= \mathbf{R}_{wb} \left( {^{b}\widetilde{\mathbf{a}}} - {^{b}\mathbf{b}_{a}} - \mathbf{n}_{a} \right) + {^{w}\mathbf{g}}, {^{b}\dot{\mathbf{b}}_{g}} = \mathbf{n}_{bg}, {^{b}\dot{\mathbf{b}}_{a}} = \mathbf{n}_{ba} \notag \\ 
        {^{w}\dot{\mathbf{g}}} &= \mathbf{0}, \quad \dot{\mathbf{R}}_{bl} = \mathbf{0}_{3 \times 3}, \quad {^{b}\dot{\mathbf{p}}_{bl}} = \mathbf{0} ,
    \end{align}
where $[ \mathbf{t} ]_{\times}$ is the skew-symmetric matrix of a 3D vector $\mathbf{t}$. 

\subsubsection{Discrete-Time Model}
Using discrete IMU measurements, we can propagate the nominal state from (\ref{eq:continuous_model}) when every IMU measurement (denoting $i$ the index) arrives \cite{FAST-LIO2}: 
    \begin{equation}
    \label{eq:discrete_model}
    \begin{aligned}
        \mathbf{x}_{i + 1} = \mathbf{x}_{i} \boxplus \left( \mathbf{f} \left( \mathbf{x}_{i}, \mathbf{u}_{i}, \mathbf{w}_{i} \right){\Delta t} \right) ,
    \end{aligned}
    \end{equation}
where $\Delta t$ is the sampling period between two adjacent IMU measurements. $\mathbf{f}$ is the derivative of the discrete dynamics model \cite{FAST-LIO2}. 
The estimated nominal state $\widehat{\mathbf{x}}$ is propagated using the above function where the noise is set to zero: 
    \begin{equation}
    \label{eq:nominal_propagation}
    \begin{aligned}
        \widehat{\mathbf{x}}_{i + 1} = \widehat{\mathbf{x}}_{i} \boxplus \left(  \mathbf{f} \left( \widehat{\mathbf{x}}_{i}, \mathbf{u}_{i}, \mathbf{0}_{12 \times 1} \right) {\Delta t}\right) .
    \end{aligned}
    \end{equation}
Retaining first-order terms from (\ref{eq:discrete_model}) and (\ref{eq:nominal_propagation}), the error state $\delta \mathbf{x}$ and the covariance $\widehat{\mathbf{P}}$ follow the linear dynamics model: 
    \begin{equation}
    \label{eq:error_state_propagation}
    \begin{aligned}
        {{\delta} \mathbf{x}_{i + 1}} &= {\mathbf{F}_{{{\delta} \mathbf{x}}_{i}}} {{\delta} \mathbf{x}_{i}} + {\mathbf{F}_{\mathbf{w}_{i}}} {\mathbf{w}_{i}} , \\ 
        \widehat{\mathbf{P}}_{i + 1} &= {\mathbf{F}_{{{\delta} \mathbf{x}}_{i}}} {\widehat{\mathbf{P}}_{i}} {\mathbf{F}_{{{\delta} \mathbf{x}}_{i}}^{\top}} + {\mathbf{F}_{\mathbf{w}_{i}}} {\mathbf{Q}_{i}} {\mathbf{F}_{\mathbf{w}_{i}}^{\top}} ,
    \end{aligned}
    \end{equation}
where ${\mathbf{F}_{{{\delta} \mathbf{x}}_{i}}}$ and ${\mathbf{F}_{\mathbf{w}_{i}}}$ are respectively the transition matrix and noise Jacobian matrix linearized at $\widehat{\mathbf{x}}_{i}$ \cite{FAST-LIO}. $\mathbf{Q}_i$ is the covariance of the noise $\mathbf{w}_i$, which can be obtained from the offline IMU calibration \cite{IMU_Allan}. The initial state $\widehat{\mathbf{x}}_{0}$ and covariance $\widehat{\mathbf{P}}_{0}$ of the current propagation are the updated state $\overline{\mathbf{x}}_{k - 1}$ and covariance $\overline{\mathbf{P}}_{k - 1}$ at the last scan-end time $t_{k -1}$.
\subsection{Motion Compensation and State Update} 

\subsubsection{Motion Compensation for Doppler Measurements} 
A generic LiDAR usually returns accumulated scans at a fixed frequency (e.g., 10 Hz), instead of returning each individual LiDAR point separately. The system is propagated from the last update at the last scan-end time $t_{k - 1}$ until the current scan arrives at time $t_k$. 
However, although a LiDAR scan arrives at a certain time, each LiDAR point in this scan is sampled at its individual discrete time. Hence, for a moving LiDAR, each point is sampled with respect to the LiDAR position at its sampling time, instead of the LiDAR position at the scan-end time. To exactly use LiDAR points in state update, we need to correct point positions in a scan by motion compensation. 

For a LiDAR point $p_j$ sampled at $t_{j}$ between two adjacent IMU sampling moments $t_{i - 1}$ and $t_{i}$, we can perform backward propagation \cite{FAST-LIO} with IMU to correct its position measurements, obtaining the corrected position $^{l_k}{\widetilde{\mathbf{p}}}_{l_k p_j}$ of $p_j$ with respect to the LiDAR position at the scan-end time $t_k$. We denote the raw position measurement of $p_j$ as $^{l_j}{\widetilde{\mathbf{p}}}_{l_j p_j}$, which is with respect to the LiDAR position at its sampling time $t_{j}$. $\check{(\cdot)}$ represents the backward propagated state at the point sampling time. The motion compensation for the point position measurement is: 
    \begin{equation}
    \begin{aligned}
    \label{eq:motion_compensation_position}
        {^{l_k}{\widetilde{\mathbf{p}}}_{l_k p_j}} = \  &{{\widehat{\mathbf{R}}}_{bl}^{\top}} \left( {{\widehat{\mathbf{R}}}_{wb_k}^{\top}} \left( {\check{\mathbf{R}}_{wb_j}} \left( {\check{\mathbf{R}}_{bl}} {^{l_j}{\widetilde{\mathbf{p}}}_{l_j p_j}} + {^{b}{\check{\mathbf{p}}}_{bl}} \right) \right. \right. \\ 
        &+ \left. \left. {^{w}{\check{\mathbf{p}}}_{wb_j}} - {^{w}{\widehat{\mathbf{p}}}_{wb_k}} \right) - {^{b}{\widehat{\mathbf{p}}}_{bl}} \right) ,
    \end{aligned}
    \end{equation}
where $( {\widehat{\mathbf{R}}}_{wb_k}, {^{w}{\widehat{\mathbf{p}}}_{wb_k}} )$ and $( {\check{\mathbf{R}}}_{wb_j}, {^{w}{\check{\mathbf{p}}}_{wb_j}} )$ are respectively system poses at the scan-end time $t_k$ and the sampling time $t_j$ of point $p_j$. This step can achieve the motion compensation for positions except for Doppler measurements. Next, we design an accurate motion compensation method for Doppler measurements, enabling the state update at $t_k$. 

Under the implicit assumption in the above position compensation that LiDAR points are static in the world frame (i.e., $^{w}\mathbf{v}_{wp_j} \equiv \mathbf{0}$), we can derive a motion compensation method for Doppler measurements. For a static point $p_j$, its Doppler velocity ${\widetilde{v}}^{d_j}_{j}$ is measured with respect to the LiDAR position and LiDAR velocity at its sampling time $t_j$ \cite{DICP}: 
    \begin{equation}
    \label{eq:doppler_j}
    \begin{aligned}
        {{\widetilde{v}}^{d_j}_{j}} = - \frac{{^{l_j}{\widetilde{\mathbf{p}}}_{l_j p_j}^{\top}}}{\left\| {^{l_j}{\widetilde{\mathbf{p}}}_{l_j p_j}} \right\|} {^{l_j}{\check{\mathbf{v}}}_{wl_j}} ,
    \end{aligned}
    \end{equation}
where $^{l_j}{\check{\mathbf{v}}}_{wl_j}$ is the LiDAR velocity with respect to the world frame and projected in the LiDAR frame at $t_j$. For $p_j$, its Doppler velocity ${\widetilde{v}}^{d_k}_{j}$, with respect to the LiDAR position and LiDAR velocity at the scan-end time $t_k$, is unequal to the raw measurement ${\widetilde{v}}^{d_j}_{j}$ at $t_j$. Similar to (\ref{eq:doppler_j}), ${\widetilde{v}}^{d_k}_{j}$ can be derived as: 
    \begin{equation}
    \label{eq:doppler_k}
    \begin{aligned}
        {{\widetilde{v}}^{d_k}_{j}} = - \frac{{^{l_k}{\widetilde{\mathbf{p}}}_{l_k p_j}^{\top}}}{\left\| {^{l_k}{\widetilde{\mathbf{p}}}_{l_k p_j}} \right\|} {^{l_k}{\widehat{\mathbf{v}}}_{wl_k}} ,
    \end{aligned}
    \end{equation}
where $^{l_k}{\widehat{\mathbf{v}}}_{wl_k}$ is the LiDAR velocity with respect to the world frame and projected in the LiDAR frame at $t_k$. To compensate the Doppler measurement from ${\widetilde{v}}^{d_j}_{j}$ to ${\widetilde{v}}^{d_k}_{j}$, we can use the forward and backward propagated states, LiDAR-IMU extrinsic parameters, and IMU measurements with the transformation between $^{l_j}{\check{\mathbf{v}}}_{wl_j}$ and $^{l_k}{\widehat{\mathbf{v}}}_{wl_k}$. Then substituting (\ref{eq:motion_compensation_position}), (\ref{eq:doppler_j}), and ${^{l}\mathbf{v}_{wl}} = {\mathbf{R}_{bl}^{\top}} ( {\mathbf{R}_{wb}^{\top}} {^{w}\mathbf{v}_{wb}} + {^{b}\bm{\omega}_{wb}} \times {^{b}\mathbf{p}_{bl}} )$ into (\ref{eq:doppler_k}): 
    \begin{align}
    \label{eq:motion_compensation_doppler}
        {{\widetilde{v}}^{d_k}_{j}} = \  &\frac{\left\| {^{l_j}{\widetilde{\mathbf{p}}}_{l_j p_j}} \right\|}{\left\| {^{l_k}{\widetilde{\mathbf{p}}}_{l_k p_j}} \right\|} {{\widetilde{v}}^{d_j}_{j}} - \frac{1}{\left\| {^{l_k}{\widetilde{\mathbf{p}}}_{l_k p_j}} \right\|} \left[ {^{w}{\check{\mathbf{p}}}_{l_j p_j}^{\top}} \left( {^{w}{\widehat{\mathbf{v}}}_{wl_k}} - {^{w}{\check{\mathbf{v}}}_{wl_j}} \right) \right. \notag \\ 
        &+ \left. {{\left( {^{w}{\check{\mathbf{p}}}_{wl_j}} - {^{w}{\widehat{\mathbf{p}}}_{wl_k}} \right)}^{\top}} {^{w}{\widehat{\mathbf{v}}}_{wl_k}} \right] ,
    \end{align}
where
    \begin{align}
    \label{eq:motion_compensation_doppler_terms}
        {^{w}{\check{\mathbf{p}}}_{l_j p_j}} &= {{\check{\mathbf{R}}}_{wb_j}} {{\check{\mathbf{R}}}_{bl}} {^{l_j}{\widetilde{\mathbf{p}}}_{l_j p_j}} , \notag \\ 
        {^{w}{\check{\mathbf{p}}}_{wl_j}} &= {^{w}{\check{\mathbf{p}}}_{wb_j}} + {{\check{\mathbf{R}}}_{wb_j}} {^{b}{\check{\mathbf{p}}}_{bl}} , \notag \\ 
        {^{w}{\widehat{\mathbf{p}}}_{wl_k}} &= {^{w}{\widehat{\mathbf{p}}}_{wb_k}} + {{\widehat{\mathbf{R}}}_{wb_k}} {^{b}{\widehat{\mathbf{p}}}_{bl}} , \notag \\ 
        {^{w}{\check{\mathbf{v}}}_{wl_j}} &= {^{w}{\check{\mathbf{v}}}_{wb_j}} + {{\check{\mathbf{R}}}_{wb_j}} \left( {^{b_j}{\check{{\bm{\omega}}}}_{wb_j}} \times {^{b}{\check{\mathbf{p}}}_{bl}} \right) , \notag \\ 
        {^{w}{\widehat{\mathbf{v}}}_{wl_k}} &= {^{w}{\widehat{\mathbf{v}}}_{wb_k}} + {{\widehat{\mathbf{R}}}_{wb_k}} \left( {^{b_k}\widehat{\bm{\omega}}_{wb_k}} \times {^{b}{\widehat{\mathbf{p}}}_{bl}} \right) .
    \end{align}
In (\ref{eq:motion_compensation_doppler_terms}), $( {{\widehat{\mathbf{R}}}_{wb_k}}, {^{w}{\widehat{\mathbf{p}}}_{wb_k}}, {^{w}{\widehat{\mathbf{v}}}_{wb_k}} )$ are forward propagated states at $t_k$, $( {{\check{\mathbf{R}}}_{wb_j}}, {^{w}{\check{\mathbf{p}}}_{wb_j}}, {^{w}{\check{\mathbf{v}}}_{wb_j}} )$ are backward propagated states at $t_j$. ${^{b_j}{\check{\bm{\omega}}}_{wb_j}}$ is the unbiased gyroscope measurement (i.e., ${^{b_j}{\check{{\bm{\omega}}}}_{wb_j}} = {^{b_j}{\widetilde{\bm{\omega}}}_j} - {^{b_j}{\check{\mathbf{b}}}_{g}}$) at $t_j$ from the IMU measurement interpolation, and ${^{b_k}\widehat{\bm{\omega}}_{wb_k}}$ is the unbiased gyroscope measurement at $t_k$. \textit{Notably, beyond motion compensation, (\ref{eq:motion_compensation_doppler}) essentially reveals the interrelation between the Doppler velocities of a static point with respect to arbitrary different LiDAR frames.} Finally, we can align all points in a scan to the scan-end time $t_k$ under a unified temporal-spatial frame, by the motion compensation for both position (\ref{eq:motion_compensation_position}) and Doppler (\ref{eq:motion_compensation_doppler}) measurements. 
    
\subsubsection{LiDAR Velocity Observation} 
In a motion-compensated scan at $t_k$, each point can provide the radial information of the LiDAR velocity in the LiDAR frame. As in (\ref{eq:doppler_k}), the Doppler velocity of a static point can be predicted using the point direction and the LiDAR velocity. The normalized direction vector ${^{l_k}{\widetilde{\mathbf{d}}}_{l_k p_j}}$ of point $p_j$ after motion compensation can be computed through: ${^{l_k}{\widetilde{\mathbf{d}}}_{l_k p_j}} = \frac{{^{l_k}{\widetilde{\mathbf{p}}}_{l_k p_j}}}{\left\| {^{l_k}{\widetilde{\mathbf{p}}}_{l_k p_j}} \right\|}$. For $n$ points, a matrix formulation can be yielded by stacking all points: 
    \begin{equation}
    \label{eq:lsq}
    \begin{aligned}
        {{\widetilde{\mathbf{D}}}_k} {^{l_k}\mathbf{v}_{wl_k}} = {{\widetilde{\mathbf{V}}}_k} ,
    \end{aligned}
    \end{equation}
where ${{\widetilde{\mathbf{D}}}_k} = {\begin{bmatrix} {\hdots} & {- {^{l_k}{\widetilde{\mathbf{d}}}_{l_k p_j}}} & {\hdots} \end{bmatrix}}^{\top}$ is a $n \times 3$ matrix and ${{\widetilde{\mathbf{V}}}_k} = {\begin{bmatrix} {\hdots} & {{\widetilde{v}}^{d_k}_{j}} & {\hdots} \end{bmatrix}}^{\top}$ is a $n \times 1$ vector. Accordingly, we can estimate a LiDAR velocity measurement at the scan-end time $t_k$ by solving a least squares problem \cite{rio}: 
    \begin{equation*}
    \begin{aligned}
        {^{l_k}{\widetilde{\mathbf{v}}}_{wl_k}} = \mathop{\arg\min} \limits_{{^{l_k}\mathbf{v}_{wl_k}}} { \left\| {{\widetilde{\mathbf{D}}}_k} {^{l_k}\mathbf{v}_{wl_k}} \! - \! {{\widetilde{\mathbf{V}}}_k} \right\| }^{2} = {{\left( {{\widetilde{\mathbf{D}}}_k^{\top}} {{\widetilde{\mathbf{D}}}_k} \right)}^{-1}} {{\widetilde{\mathbf{D}}}_k^{\top}} {{\widetilde{\mathbf{V}}}_k} .
    \end{aligned}
    \end{equation*}
Meanwhile, the covariance of the estimated LiDAR velocity, ${\mathbf{R}}_{\mathbf{v}}$, can be computed from the estimation residual \cite{rio}, serving as the observation covariance in the state update. 

For an FMCW Doppler LiDAR scan containing tens of thousands of points, directly solving the above problem is computationally burdensome and error-prone, due to potential dynamic points. Thus, a 3-point Random Sample and Consensus (RANSAC) method \cite{rio} is applied to efficiently obtain a robust solution. For example, setting the outlier probability to 30\% and the desired success probability to 99\%, the iteration number is 11, which leads to a high efficiency. Besides, to prevent the solution from falling into the null-space of (\ref{eq:lsq}) (i.e., ${{\widetilde{\mathbf{D}}}_k} {^{l_k}\mathbf{v}_{wl_k}} = {\mathbf{0}_{n \times 1}}$) and improve the ratio of valid samples, the sampling probability is weighted by the absolute cosine of the angle ${{\phi}_{{{\widehat{\mathbf{v}}}_{l}} {\widetilde{\mathbf{d}}_{j}}}}$ between the point direction and the forward propagated LiDAR velocity in the LiDAR frame: 
    \begin{equation}
    \label{eq:cos_weight}
    \begin{aligned}
        \left\| \cos \left( {{\phi}_{{{\widehat{\mathbf{v}}}_{l}} {\widetilde{\mathbf{d}}_{j}}}} \right) \right\| = \frac{ \bigl\| \langle {^{l_k}{\widetilde{\mathbf{d}}}_{l_k p_j}} , {^{l_k}{\widehat{\mathbf{v}}}_{wl_k}} \rangle \bigr\| }{ \bigl\| {^{l_k}{\widetilde{\mathbf{d}}}_{l_k p_j}} \bigr\| \bigl\| {^{l_k}{\widehat{\mathbf{v}}}_{wl_k}} \bigr\| } ,
    \end{aligned}
    \end{equation}
where ${^{l_k}{\widehat{\mathbf{v}}}_{wl_k}} = {\left( {{\widehat{\mathbf{R}}}_{wb_k}} {{\widehat{\mathbf{R}}}_{bl}} \right)}^{\top} {^{w}{\widehat{\mathbf{v}}}_{wl_k}}$ is the forward propagated LiDAR velocity in the LiDAR frame at the scan-end time $t_k$, and ${^{w}{\widehat{\mathbf{v}}}_{wl_k}}$ can be computed from (\ref{eq:motion_compensation_doppler_terms}).

The above estimated measurement of LiDAR velocity provides a correspondence-free observation for the system state. The LiDAR velocity observation model ${{\mathbf{h}}_{\mathbf{v}} {\left( {\mathbf{x}}, {^{l}\mathbf{n}_{\mathbf{v}}} \right)}}$ is: 
    \begin{equation*}
    \begin{aligned}
        {^{l}{{\widetilde{\mathbf{v}}}}_{wl}} &= {{\mathbf{h}}_{\mathbf{v}} {\left( {\mathbf{x}}, {^{l}\mathbf{n}_{\mathbf{v}}} \right)}} \\ 
        &\triangleq {\mathbf{R}_{bl}^{\top}} \left( {\mathbf{R}_{wb}^{\top}} {^{w}\mathbf{v}_{wb}} + \left( {^{b}\widetilde{\bm{\omega}}} - {^{b}\mathbf{b}_{g}} - \mathbf{n}_{g} \right) \times {^{b}\mathbf{p}_{bl}} \right) + {^{l}\mathbf{n}_{\mathbf{v}}} ,
    \end{aligned}
    \end{equation*}
where the covariance of ${^{l}\mathbf{n}_{\mathbf{v}}}$ is the aforementioned ${\mathbf{R}}_{\mathbf{v}}$. To reach the optimal estimate at $t_k$, the LiDAR velocity residual ${\mathbf{r}_{\mathbf{v}_k}} = {^{l_k}{\widetilde{\mathbf{v}}}_{wl_k}} - {^{l_k}{\widehat{\mathbf{v}}}_{wl_k}}$ can be linearized as: 
    \begin{equation}
    \label{eq:lidar_vel_residual_linear}
    \begin{aligned}
        {\mathbf{r}_{\mathbf{v}_k}} = {\mathbf{h}}_{\mathbf{v}} {\left( {\mathbf{x}_k}, {\mathbf{n}_{\mathbf{v}}} \right)} - {\mathbf{h}}_{\mathbf{v}} {\left( {{\widehat{\mathbf{x}}}_k}, {\mathbf{0}} \right)} \approx {\mathbf{H}_{\mathbf{v}_k}} {{\delta} \mathbf{x}_{k}} + {^{l}\mathbf{n}_{\mathbf{v}}} .
    \end{aligned}
    \end{equation}
The observation Jacobian matrix ${\mathbf{H}_{\mathbf{v}_k}} \in {\mathbb{R}^{3 \times 24}}$ is: 
    \begin{equation*}
    \setlength{\arraycolsep}{4.0pt}
    \begin{aligned}
        {\mathbf{H}_{\mathbf{v}_k}} = {\begin{bmatrix} {\mathbf{H}_{\delta {\bm{\theta}}_{k}}} & {\mathbf{0}_{3 \times 3}} & {\mathbf{H}_{{{\delta} \mathbf{v}_{k}}}} & {\mathbf{H}_{{{\delta} \mathbf{b}_{g k}}}} & {\mathbf{0}_{3 \times 6}} & {\mathbf{H}_{\delta {\bm{\theta}}_{bl}}} & {\mathbf{H}_{{\delta} {\mathbf{p}}_{bl}}} \end{bmatrix}} .
    \end{aligned}
    \end{equation*}
The different block components of ${\mathbf{H}_{\mathbf{v}_k}}$ are: 
    \begin{align*}
        &{\mathbf{H}_{\delta {\bm{\theta}}_{k}}} = {{\widehat{\mathbf{R}}}_{bl}^{\top}} { [ {{\widehat{\mathbf{R}}}_{wb_k}^{\top}} {^{w}{\widehat{\mathbf{v}}}_{wb_k}} ] }_{\times}, \  {\mathbf{H}_{{{\delta} \mathbf{v}_{k}}}} = { ( {{\widehat{\mathbf{R}}}_{wb_k}} {{\widehat{\mathbf{R}}}_{bl}} ) }^{\top} \\ 
        &{\mathbf{H}_{{{\delta} \mathbf{b}_{g k}}}} = {{\widehat{\mathbf{R}}}_{bl}^{\top}} { [ {^{b}{\widehat{\mathbf{p}}}_{bl}} ] }_{\times}, \  {\mathbf{H}_{{\delta} {\mathbf{p}}_{bl}}} = {{\widehat{\mathbf{R}}}_{bl}^{\top}} { [ {^{b_k}{\widetilde{\bm{\omega}}}_k} - {^{b_k}{\widehat{\mathbf{b}}}_{g}} ] }_{\times} \\ 
        &{\mathbf{H}_{\delta {\bm{\theta}}_{bl}}} = { [ {{\widehat{\mathbf{R}}}_{bl}^{\top}} { ( {{\widehat{\mathbf{R}}}_{wb_k}^{\top}} {^{w}{\widehat{\mathbf{v}}}_{wb_k}} + ( {^{b_k}{\widetilde{\bm{\omega}}}_k} - {^{b_k}{\widehat{\mathbf{b}}}_{g}} ) \times {^{b}{\widehat{\mathbf{p}}}_{bl}} ) } ] }_{\times} .
    \end{align*}
On account of the low dimension of 3D LiDAR velocity observation ($\text{dim}({\mathbf{r}_{\mathbf{v}_k}}) \ll \text{dim}(\mathcal{M})$), the conventional formulation of Kalman gain instead of (\ref{eq:kalman_gain_geo}) is applied: 
    \begin{equation*}
    \begin{aligned}
        {\mathbf{K}_{\mathbf{v}_k}} = {\widehat{\mathbf{P}}_{k}} {\mathbf{H}_{\mathbf{v}_k}^{\top}} { \left( {\mathbf{H}_{\mathbf{v}_k}} {\widehat{\mathbf{P}}_{k}} {\mathbf{H}_{\mathbf{v}_k}^{\top}} + {\mathbf{R}}_{\mathbf{v}} \right) }^{-1} .
    \end{aligned}
    \end{equation*}
The updated state ${\overline{\mathbf{x}}}_{{\mathbf{v}}_k}$ and covariance ${\overline{\mathbf{P}}}_{{\mathbf{v}}_k}$ using the LiDAR velocity observation are respectively: 
    \begin{align}
    \label{eq:lidar_vel_update}
        {\overline{\mathbf{x}}}_{{\mathbf{v}}_k} &= {{\widehat{\mathbf{x}}}_{k}} \boxplus {\overline{{\delta} \mathbf{x}}_{\mathbf{v}_k}} = {{\widehat{\mathbf{x}}}_{k}} \boxplus \left( {\mathbf{K}_{\mathbf{v}_k}} {\mathbf{r}_{\mathbf{v}_k}} \right) , \notag \\ 
        {\overline{\mathbf{P}}}_{{\mathbf{v}}_k} &= {\mathbf{L}_{\mathbf{v}_k}} \left( {{\mathbf{I}}_{24 \times 24}} - {\mathbf{K}_{\mathbf{v}_k}} {\mathbf{H}_{\mathbf{v}_k}} \right) {\widehat{\mathbf{P}}_{k}} {\mathbf{L}_{\mathbf{v}_k}^{\top}} ,
    \end{align}
where ${\mathbf{L}_{\mathbf{v}_k}}$ is the projection matrix between tangent spaces arising from the on-manifold increment ${\overline{{\delta} \mathbf{x}}_{\mathbf{v}_k}}$ \cite{IKFoM-arXiv}. 

    \begin{figure}[!t]
       \centering
       \includegraphics[width=0.47\textwidth]{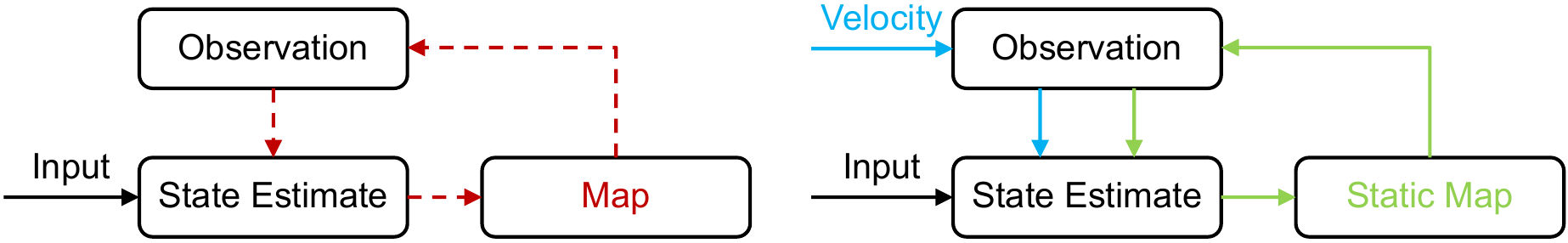}
       \caption{Illustration of the feedback loop between the state estimate and the online map.} 
       \label{fig:feedback_loop}
    \end{figure}
    
\subsubsection{LiDAR Geometric Observation}
Apart from the instant correspondence-free Doppler-aided observation, the geometric association from LiDAR points can also provide observations. Here we use the point-plane distance model \cite{FAST-LIO2} \cite{LOAM}: 
    \begin{equation}
    \label{eq:lidar_geo_obs}
    \begin{aligned}
        {0} &= {{h}_{g_j} {\left( {\mathbf{x}}, {^{l}\mathbf{n}_{p_j}} \right)}} \\ 
        &\triangleq {^{w}{\mathbf{n}}_{plane_j}^{\top}} \left( {\mathbf{T}_{wb}} {\mathbf{T}_{bl}} \left( {^{l}{\widetilde{\mathbf{p}}}_{l p_j}} - {^{l}\mathbf{n}_{p_j}} \right) - {^{w}{\widetilde{\mathbf{q}}}_{w q_j}} \right) ,
    \end{aligned}
    \end{equation}
where $\mathbf{T}_{wb}, \mathbf{T}_{bl} \in SE(3)$ are respectively the system pose and extrinsic parameters, ${^{l}\mathbf{n}_{p_j}}$ is the point position noise, ${^{w}{\mathbf{n}}_{plane_j}}$ is the unit normal vector of the nearest neighbor plane of point $p_j$ in the world frame, and ${^{w}{\widetilde{\mathbf{q}}}_{w q_j}}$ is a point in that plane projected in the world frame. 
    
In IESEKF, the state updated by the LiDAR velocity observation give a prior Gaussian distribution in the tangent space at ${\overline{\mathbf{x}}}_{{\mathbf{v}}_k}$, but the optimal estimate is derived in the tangent space at $\widehat{\mathbf{x}}_{k}^{\kappa}$, i.e., the state of $\kappa$-th update iteration (equals the updated state after ($\kappa - 1$)-th update iteration):   
    \begin{equation*}
    \begin{aligned}
        {\widehat{\mathbf{x}}_{k}^{\kappa}} = {\overline{\mathbf{x}}_{k}^{\kappa - 1}} = {\widehat{\mathbf{x}}_{k}^{\kappa - 1}} \boxplus {\overline{\delta \mathbf{x}}_{k}^{\kappa - 1}}, \  {\widehat{\mathbf{x}}_{k}^{0}} = {\overline{\mathbf{x}}}_{{\mathbf{v}}_k} .
    \end{aligned}
    \end{equation*}
With ${{\delta} \mathbf{x}_{k}} = {{\mathbf{x}}_{k}} \boxminus {\overline{\mathbf{x}}}_{{\mathbf{v}}_k} \sim \mathcal{N} \left( \mathbf{0}_{24 \times 1} , {\overline{\mathbf{P}}}_{{\mathbf{v}}_k} \right)$, the prior distribution projected in the tangent space at $\widehat{\mathbf{x}}_{k}^{\kappa}$ imposing a prior error-state distribution in the $\kappa$-th update iteration: 
    \begin{equation*}
    \begin{aligned}
        {{\delta} \mathbf{x}_{k}^{\kappa}} \sim \mathcal{N} \left( {- \mathbf{J}_{k}^{\kappa} \left( {\widehat{\mathbf{x}}_{k}^{\kappa}} \boxminus {\overline{\mathbf{x}}}_{{\mathbf{v}}_k} \right)} , {\mathbf{J}_{k}^{\kappa} {\overline{\mathbf{P}}}_{{\mathbf{v}}_k} \mathbf{J}_{k}^{\kappa \top}} \right) ,
    \end{aligned}
    \end{equation*}
where ${\mathbf{J}_{k}^{\kappa}}$ is the tangent projection matrix because of the on-manifold update step $\left( {\widehat{\mathbf{x}}_{k}^{\kappa}} \boxminus {\overline{\mathbf{x}}}_{{\mathbf{v}}_k} \right)$, similar to (\ref{eq:lidar_vel_update}). The LiDAR geometric observation residual of point $p_j$ at $t_k$ is: 
    \begin{equation*}
    \begin{aligned}
        \mathbf{r}_{{g_k}_j}^{\kappa} = 0 - {{h}_{g_j} {\left( {{\widehat{\mathbf{x}}}_k^{\kappa}}, \mathbf{0} \right)}} = - {{h}_{g_j} {\left( {{\widehat{\mathbf{x}}}_k^{\kappa}}, \mathbf{0} \right)}} .
    \end{aligned}
    \end{equation*}
Then, fusing the prior distribution and the likelihood observation distribution from all observations, an equivalent maximum a-posteriori estimate (MAP) can be obtained \cite{IKFoM-arXiv}. 

Therefore, the iterated state update follows:  
    \begin{align}
    \label{eq:kalman_gain_geo}
        {\mathbf{K}_{g_k}^{\kappa}} &= { \left( {\mathbf{H}_{g_k}^{\kappa \top}} {\mathbf{R}_{g_k}^{-1}} {\mathbf{H}_{g_k}^{\kappa}} + {\left( {\mathbf{J}_{k}^{\kappa} {\overline{\mathbf{P}}}_{{\mathbf{v}}_k} \mathbf{J}_{k}^{\kappa \top}} \right)}^{-1} \right) }^{-1} {\mathbf{H}_{g_k}^{\kappa \top}} {\mathbf{R}_{g_k}^{-1}} , \notag \\ 
        {\overline{\delta \mathbf{x}}_{k}^{\kappa}} &= {\mathbf{K}_{g_k}^{\kappa}} {\mathbf{r}_{g_k}^{\kappa}} + \left( {\mathbf{K}_{g_k}^{\kappa}} {\mathbf{H}_{g_k}^{\kappa}} - {\mathbf{I}_{24 \times 24}} \right) {\mathbf{J}_{k}^{\kappa} \left( {\widehat{\mathbf{x}}_{k}^{\kappa}} \boxminus {\overline{\mathbf{x}}}_{{\mathbf{v}}_k} \right)} , \notag \\ 
        {\widehat{\mathbf{x}}_{k}^{\kappa + 1}} &= {\widehat{\mathbf{x}}_{k}^{\kappa}} \boxplus {\overline{\delta \mathbf{x}}_{k}^{\kappa}} ,
    \end{align}
where ${\mathbf{r}_{g_k}^{\kappa}}$, ${\mathbf{H}_{g_k}^{\kappa}}$, and ${\mathbf{R}_{g_k}}$ are formulated by stacking matrices of all valid points. Finally, after the convergence of IESEKF, the optimally updated state and covariance are: 
    \begin{equation}
    \label{eq:optimal_state_cov}
    \begin{aligned}
        {\overline{\mathbf{x}}_{k}} &= {\overline{\mathbf{x}}_{g_k}} = {\widehat{\mathbf{x}}_{k}^{\kappa + 1}} , \\ 
        {\overline{\mathbf{P}}}_{k} &= {\overline{\mathbf{P}}}_{g_k} = {\mathbf{L}_{g_k}} \left( {{\mathbf{I}}_{24 \times 24}} - {\mathbf{K}_{g_k}^{\kappa}} {\mathbf{H}_{g_k}^{\kappa}} \right) {\left( {\mathbf{J}_{k}^{\kappa} {\overline{\mathbf{P}}}_{{\mathbf{v}}_k} \mathbf{J}_{k}^{\kappa \top}} \right)} {\mathbf{L}_{g_k}^{\top}} .
    \end{aligned}
    \end{equation}
In (\ref{eq:optimal_state_cov}), ${\overline{\mathbf{x}}_{g_k}}$ and ${\overline{\mathbf{P}}}_{g_k}$ are the updated state and covariance using the LiDAR geometric observation, ${\overline{\mathbf{x}}_{k}}$ and ${\overline{\mathbf{P}}}_{k}$ are the optimal state estimate and covariance after both LiDAR velocity and geometric update at the scan-end time $t_k$. 
\subsection{Dynamic Point Removal and Static Mapping} \label{sec:dyn_removal_static_mapping} 

The LiDAR geometric observation in (\ref{eq:lidar_geo_obs}) depends on the online established map. Dynamic points in the LiDAR scan can break the static world assumption and introduce significant errors. Importantly, dynamic points can contaminate the online map, which is prone to provide erroneous geometric correspondences. This dependence creates a loop between the state estimate and the map that can destabilize the system, as shown in the left of Fig. \ref{fig:feedback_loop}. Specifically, if the state estimate is disturbed by dynamic points, the online map will be inaccurate and chaotic, and the erroneous correspondences from dynamic points will amplify the estimation error. This positive feedback loop of the estimation error may lead to divergence even failure of the state estimator. 

Fortunately, the above problem in traditional LiDAR systems can be solved directly and efficiently thanks to the inherent Doppler measurements from FMCW Doppler LiDARs. After the state update in (\ref{eq:lidar_vel_update}), we can predict a Doppler velocity ${{\widetilde{v}}^{d_k}_{j, pred}}$ for point $p_j$. If the discrepancy between the prediction and the measurement exceeds a given threshold, the point is removed to prevent it from being added to the LiDAR geometric observation and the online map. The method is similar to the outlier rejection mentioned in \cite{DICP} but the threshold $\delta v^{d}_{thr}$ is adjustable according to the angle ${{\phi}_{{{\overline{\mathbf{v}}}_{l}} {\widetilde{\mathbf{d}}_{j}}}}$ between the normalized direction of a point and the LiDAR velocity from the updated state in (\ref{eq:lidar_vel_update}). The condition that the point $p_j$ is regarded as a dynamic point is: 
    \begin{equation}
    \label{eq:dyn_removal_thr}
    \begin{aligned}
        {\left\| {{\widetilde{v}}^{d_k}_{j, pred}} - {{\widetilde{v}}^{d_k}_{j}} \right\|} > {\delta v^{d}_{thr}} {\left\| \cos \left( {{\phi}_{{{\overline{\mathbf{v}}}_{l}} {\widetilde{\mathbf{d}}_{j}}}} \right) \right\|} ,
    \end{aligned}
    \end{equation}
where the cosine is in (\ref{eq:cos_weight}) but uses the updated state ${^{l_k}{\overline{\mathbf{v}}}_{wl_k}}$.

Significantly, as shown in (\ref{eq:cos_weight}) and (\ref{eq:dyn_removal_thr}), this cosine reveals a crucial physical insight: the effectiveness and the signal-to-noise ratio of a LiDAR point in estimating the 3D linear LiDAR velocity depend on the angle (relative direction) between the point direction and the LiDAR velocity projected in the LiDAR frame (i.e., ${^{l}{\mathbf{v}}_{wl}}$), rather than directly on the LiDAR velocity projected in the world frame (i.e., ${^{w}{\mathbf{v}}_{wl}}$) or the Field-of-View (FoV) of an FMCW Doppler LiDAR. Meanwhile, this cosine represents the radial nature of the Doppler effect. 

Consequently, only almost static points are fed into the LiDAR geometric observation and the map, which can further improve the accuracy and robustness of the online system. FMCW-LIO can eventually maintain a neat and static map. 

Meanwhile, the Doppler-aided correspondence-free observations can partially alleviate the drawback of the aforementioned positive error feedback loop. This observation model reduces the reliance on historical associations from the online map, as shown in the right of Fig. \ref{fig:feedback_loop}. 
    \begin{figure}[!t]
       \centering
       \includegraphics[width=0.47\textwidth]{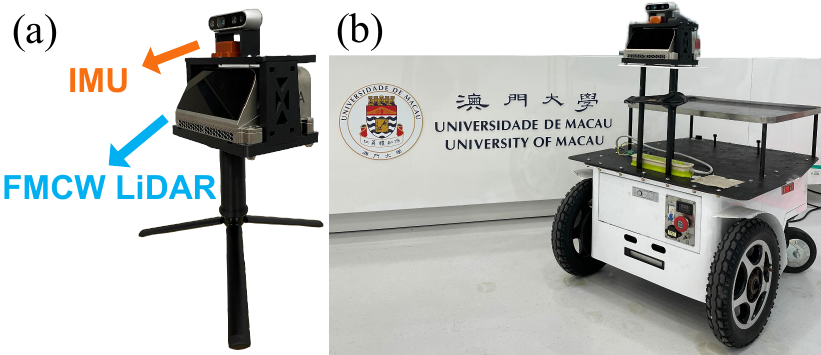}
       \caption{The sensor suite on the (a) handheld and the (b) wheeled platforms.} 
       \label{fig:sensor_suite}
    \end{figure}


\section{Experimental Results} \label{sec:experimental_results}

\subsection{FMCW Doppler LiDAR-Inertial Data Collection} 

It is necessary to collect the FMCW Doppler LiDAR-inertial data in the real world to validate the proposed method, due to the absence of public FMCW Doppler LiDAR-inertial datasets including raw Doppler measurements, as mentioned in Section \ref{subsec:fmcw_related_work}. We use an Aeva Aeries II \cite{aeva2023} FMCW Doppler LiDAR and an Xsens MTi-G-710 IMU to assemble our sensor suite. This sensor suite can be mounted on handheld or wheeled platforms, as in Fig. \ref{fig:sensor_suite}. The LiDAR FoV is 120$^{\circ}$$\times$28.8$^{\circ}$. The LiDAR returns 10 Hz scans including position and raw Doppler measurements. The frequency of IMU measurements is 200 Hz. The computing device is the Intel NUC with an i7-1165G7 CPU. 

\begin{figure}[!t]
       \centering
       \includegraphics[width=0.47\textwidth]{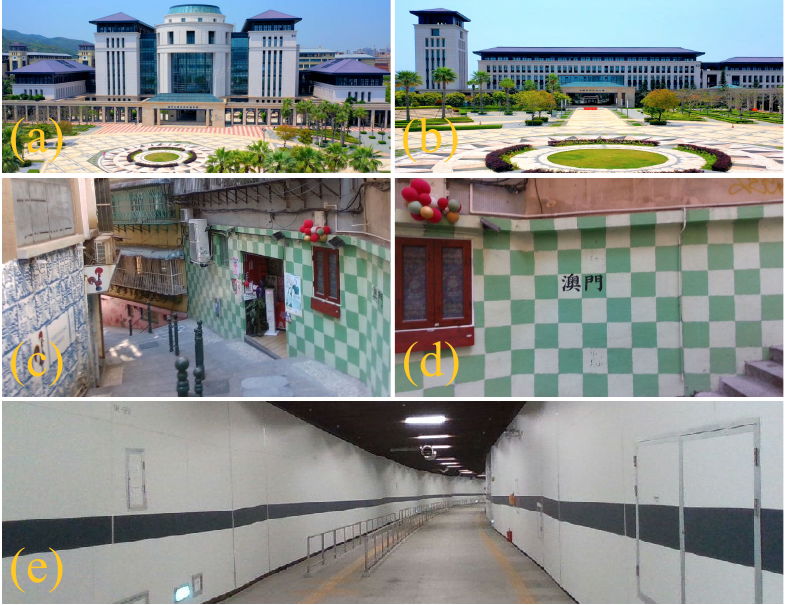}
       \caption{The scenes of the collected dataset. (a) and (b) are campus scenes. (c) and (d) are narrow street scenes. (e) is the tunnel scene.} 
       \label{fig:campus_narrow_street_tunnel_img}
    \end{figure}

The data description is shown in Table \ref{tab:table_dataset_description}. For the handheld platform, the sensor suite is held by a walking person with persistent shaking and swing. The speed of the handheld platform is slow (about 1.5 m/s). For the wheeled platform, the sensor suite is mounted on a vehicle and the moving speed is relatively fast (up to 7.0 m/s). To comprehensively evaluate different algorithms, diverse conditions (structured vs. structure-degenerated scenes, wide vs. narrow scenes, slow vs. fast speeds) are considered, as shown in Fig. \ref{fig:campus_narrow_street_tunnel_img}. In \textit{campus\_01} and \textit{campus\_02}, we traverse the campus of the University of Macau and the scene is wide enough to extract and match geometric features for general LIO systems. On the contrary, \textit{street\_01} and \textit{street\_02} are collected in narrow streets in downtown Macao, thus the scene is considerably narrow leading to significant challenges for LIO systems. Furthermore, to validate the Doppler-aided observation, four structure-degenerated sequences are collected in a subaqueous pedestrian tunnel. The tunnel is S-shaped and the one-side length is about 850 m. Since the dataset is challenging and the GNSS accuracy is highly unreliable in these narrow, among-building, or indoor scenes, we choose the trajectory optimized by offline post-processing as the ground-truth reference. We utilize a state-of-the-art loop closure \cite{wang2020lidar_iris} and the pose graph optimization \cite{dellaert2012GTSAM} to generate the ground truth. Notably, all post-processed ground-truth trajectories are meticulously examined to ensure alignment with the real world. Besides, we use two metrics, absolute translational error (RMSE) and end-to-end error (E2E) \cite{FAST-LIO2}, to evaluate algorithms. Therefore, the starting and ending positions of all sequences are confirmed to be exactly coincident to enable the end-to-end evaluation. 

For the thorough evaluation and ablation study of different algorithms and modules, we compare FMCW-LIO with conventional mainstream LIO systems under diverse framework designs, i.e., LINS \cite{LINS}, LIO-SAM \cite{LIO-SAM}, DLIO \cite{DLIO}, and FAST-LIO2 \cite{FAST-LIO2}. We also evaluate the recently released Doppler-aided LO, STEAM-DICP \cite{Picking_up_Speed}, which can serve as an ablation study to validate the role of IMU on handheld or small and medium-scale robot platforms with narrow sensing FoV, narrow workspaces, and high-frequency dynamic motions. We tune the parameters of all algorithms to achieve their optimal accuracy and use these tuned parameters in all experiments. In FMCW-LIO, we use \textit{ikd-Tree} \cite{FAST-LIO2} as the map structure, and the local map size is set to 1000 m. We also test two variants of FMCW-LIO to perform ablation studies and validate the effectiveness of two modules, including 1) the variant without the Doppler compensation (w/o DC) and 2) the variant without the dynamic point removal (w/o DR).
\begin{table}[!t] 
    \setlength
    \tabcolsep{3.8pt}
    \caption{Description of the Collected Dataset} 
    \label{tab:table_dataset_description} 
    \begin{center} 
        \begin{tabular}{cccccc} 
            \toprule 
            Sequence & Platform & Speed & Scene Type & Length ($m$) & Duration ($s$) \\ 
            \midrule
            \textit{campus\_01} & Handheld & Slow & Wide & 425 & 403 \\ 
            \textit{campus\_02} & Wheeled & Fast & Wide & 454 & 172 \\ 
            \textit{street\_01} & Handheld & Slow & Narrow & 182 & 204 \\ 
            \textit{street\_02} & Handheld & Slow & Narrow & 186 & 223 \\ 
            \textit{tunnel\_01} & Handheld & Slow & Degenerated & 1129 & 740 \\ 
            \textit{tunnel\_02} & Wheeled & Fast & Degenerated & 232 & 90 \\ 
            \textit{tunnel\_03} & Handheld & Slow & Degenerated & 283 & 179 \\ 
            \textit{tunnel\_04} & Wheeled & Fast & Degenerated & 1712 & 368 \\ 
            \bottomrule
        \end{tabular}
    \end{center}
\end{table}
\begin{table*}[!t] 
    \setlength
    \tabcolsep{4.0pt}
    \caption{Absolute Translational Errors (RMSE, Meters) and End-to-End Errors (E2E, Meters) on the Collected Dataset} 
    \label{tab:table_results_double_col} 
    \begin{center} 
        \begin{tabular}{ccccccccccccccccc} 
            \toprule
            \multirow{2}{*}{Method/Sequence} & \multicolumn{2}{c}{\textit{campus\_01}} & \multicolumn{2}{c}{\textit{campus\_02}} & \multicolumn{2}{c}{\textit{street\_01}} & \multicolumn{2}{c}{\textit{street\_02}} & \multicolumn{2}{c}{\textit{tunnel\_01}} & \multicolumn{2}{c}{\textit{tunnel\_02}} & \multicolumn{2}{c}{\textit{tunnel\_03}} & \multicolumn{2}{c}{\textit{tunnel\_04}} \\ 
                            \cmidrule(lr){2-3} \cmidrule(lr){4-5} \cmidrule(lr){6-7} \cmidrule(lr){8-9} \cmidrule(lr){10-11} \cmidrule(lr){12-13} \cmidrule(lr){14-15} \cmidrule(lr){16-17}
                            & RMSE & E2E & RMSE & E2E & RMSE & E2E & RMSE & E2E & RMSE & E2E & RMSE & E2E & RMSE & E2E & RMSE & E2E \\ 
            \midrule
            LINS & 3.18 & 34.36 & 4.24 & 38.98 & $-$ & $-$ & $-$ & $-$ & $-$ & $-$ & $-$ & $-$ & $-$ & $-$ & $-$ & $-$ \\ 
            LIO-SAM & 1.24 & 4.29 & 4.89 & 3.16 & $\times$ & $\times$ & 0.23 & 0.18 & $\times$ & $\times$ & 2.11 & 0.10 & $\times$ & $\times$ & $\times$ & $\times$ \\ 
            DLIO & 1.11 & 1.51 & 3.44 & 2.53 & $\times$ & $\times$ & 2.00 & 7.96 & $\times$ & $\times$ & 5.61 & 11.87 & $\times$ & $\times$ & $\times$ & $\times$ \\ 
            FAST-LIO2 & 0.84 & 1.08 & 0.54 & \textbf{0.07} & 11.74 & 4.33 & 0.19 & 0.05 & $\times$ & $\times$ & 14.99 & 17.50 & 46.11 & 99.83 & 60.09 & 23.84 \\ 
            STEAM-DICP & 1.23 & 1.49 & 1.76 & $\times$ & $\times$ & $\times$ & $\times$ & $\times$ & 35.16 & $\times$ & 4.39 & $\times$ & 0.81 & $\times$ & 49.32 & $\times$ \\ 
            \midrule 
            FMCW-LIO (w/o DC) & 0.11 & 0.14 & 0.87 & 0.31 & 1.14 & 0.23 & 0.20 & 0.08 & 0.96 & 0.52 & 1.94 & 1.65 & 0.59 & 0.32 & 2.04 & 0.86 \\ 
            FMCW-LIO (w/o DR) & 0.13 & 0.13 & 0.16 & 0.09 & 2.73 & 0.72 & 1.20 & 0.35 & 0.48 & 0.20 & 0.12 & 0.06 & 0.28 & \textbf{0.05} & \textbf{0.11} & 0.06 \\ 
            FMCW-LIO & \textbf{0.08} & \textbf{0.12} & \textbf{0.13} & \textbf{0.07} & \textbf{0.19} & \textbf{0.11} & \textbf{0.18} & \textbf{0.04} & \textbf{0.47} & \textbf{0.12} & \textbf{0.07} & \textbf{0.03} & \textbf{0.27} & \textbf{0.05} & 0.13 & \textbf{0.02} \\ 
            \bottomrule
        \end{tabular}
    \end{center}
        \footnotesize{
            $-$ As LINS is designed for spinning LiDARs and ground vehicles, its applicability on our LiDAR and platforms is highly challenging. \\ 
            $\times$ The algorithm failed or severely drifted in the corresponding sequence. 
        }
\end{table*}
\subsection{Experiments in Structured Environments}

The results in four structured scenes are in the left four columns of Table \ref{tab:table_results_double_col} and the loop closure of LIO-SAM is disabled for fair comparisons. FMCW-LIO achieves accuracy comparable to FAST-LIO2 in \textit{campus} scenes. However, in \textit{street\_01}, the narrow scene and intense on-stairs motions (as shown in Fig. \ref{fig:campus_narrow_street_tunnel_img}(c)) result in failures or large drifts of other algorithms, but FMCW-LIO can achieve remarkably low errors. The reason is that, in \textit{street\_01}, the sensor suite faces a feature-less wall within an extremely narrow scene which leads to erroneous or invalid geometric observations (as in Fig. \ref{fig:campus_narrow_street_tunnel_img}(d)). Nevertheless, FMCW-LIO can use the Doppler-aided LiDAR velocity observation that is independent of the geometric features of environments. Meanwhile, the LiDAR velocity observation can perform the Zero-Velocity Update (ZUPT) when the platform is steady when facing the wall or offer $z$-axis velocity when a large motion in the $z$ direction occurs. As a result, the estimated trajectory of FMCW-LIO can exactly return to the starting position. 

\begin{figure}[!t]
       \centering
       \includegraphics[width=0.47\textwidth]{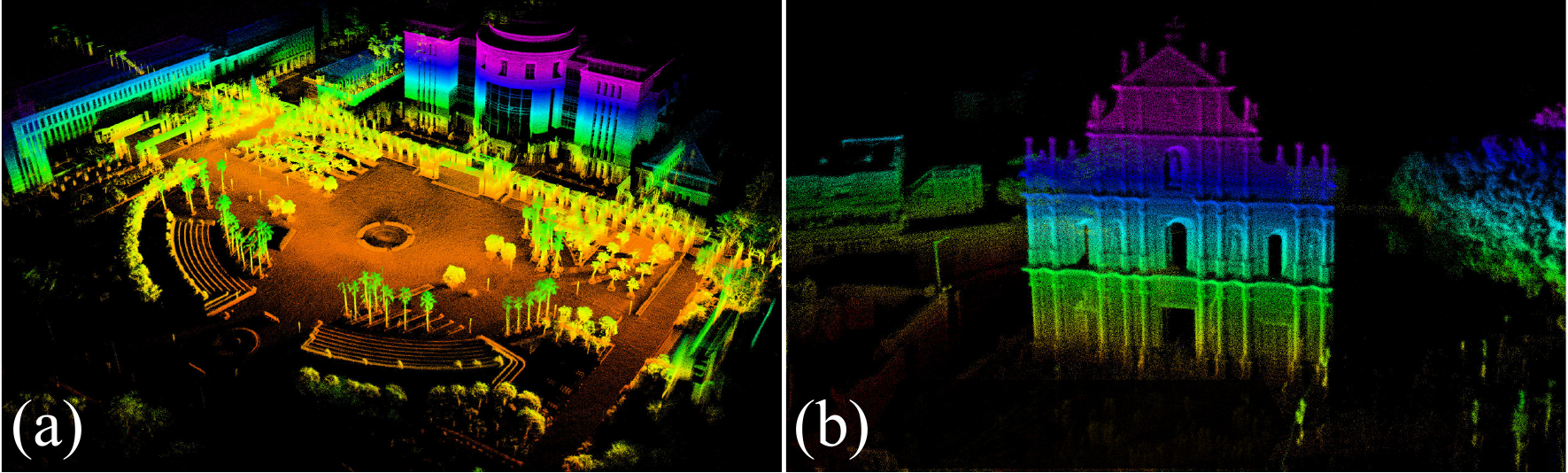}
       \caption{Mapping in structured scenes. (a) Campus scene. (b) Street scene.} 
       \label{fig:structured_mapping}
    \end{figure}

In addition, due to the presence of numerous moving pedestrians and cars in \textit{street} sequences, the variant (w/o DR) exhibits larger localization errors. Conversely, the complete FMCW-LIO and the variant (w/o DC) achieve accurate estimation performance. In structured sequences, the Doppler compensation effect is not significant due to the slow-moving speeds of both handheld and wheeled platforms operating within the campus or street blocks. Hence, the variant (w/o DC) achieves comparable accuracy with the complete FMCW-LIO. The mapping results of FMCW-LIO in structured scenes are shown in Fig. \ref{fig:structured_mapping} and the average running time per scan of algorithms is listed in Table \ref{tab:table_avg_running_time}. FMCW-LIO can achieve real-time and efficient performance that is comparable to that of FAST-LIO2. Online demonstrations can be found in the video.

Furthermore, we observe that STEAM-DICP demonstrates estimation accuracy comparable to other LIO methods in \textit{campus\_01}. However, in other sequences, it fails when encountering dynamic motions, fast rotations, or narrow scenes due to the lack of keypoints. In \textit{campus\_02}, STEAM-DICP can successfully run up to the halfway point, until rapid rotations occur. Thus, we only calculate the RMSE of the segments where it can successfully run without system failures. Therefore, it can be observed that IMU is essential for handheld or robotic platforms that frequently encounter narrow workspaces and highly dynamic motions.
\subsection{Experiments in Structure-Degenerated Environments}

\begin{figure}[!t]
       \centering
       \includegraphics[width=0.47\textwidth]{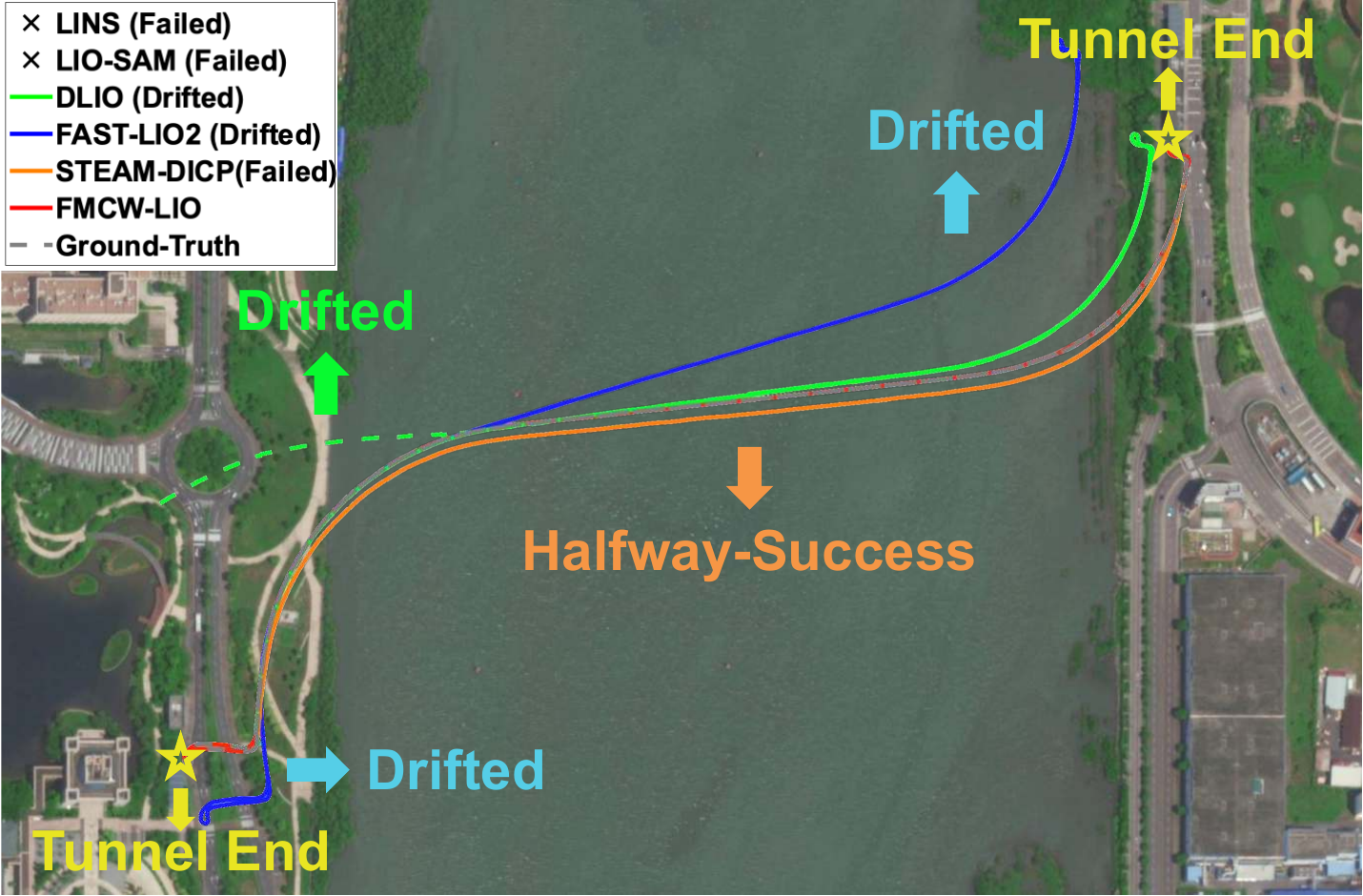}
       \caption{Estimated trajectories overlaid with the satellite map.} 
       \label{fig:traj_plot}
    \end{figure}

To further validate the Doppler-aided observation model, we evaluate algorithms on four \textit{tunnel} sequences targeting geometric structure-degenerated scenes. These sequences have diverse distributions in length and moving speed. The results are shown in the right four columns of Table \ref{tab:table_results_double_col}. As expected, all conventional LIO systems that depend on geometric observations fail in most \textit{tunnel} sequences. FMCW-LIO, however, accomplishes incredibly low errors, accurately returning to starting points. In the shortest sequence \textit{tunnel\_02}, LIO-SAM can achieve accurate localization with a factor graph framework which can maintain relatively long-term constraints in this short length. Although DLIO and FAST-LIO2 do not fail, they fail to achieve accurate localization, generating chaotic, inconsistent and drifted tunnel maps, as in Fig. \ref{fig:system_demo}(a)(c). In the longest sequence \textit{tunnel\_04}, FMCW-LIO can perfectly achieve precise localization (2 cm end-to-end error over 1.7 km traversed round-trip distance) through the whole tunnel. The Doppler-aided observation provides a direct observation of the LiDAR velocity state. However, in long tunnel sequences, the factor graph framework in LIO-SAM that only relies on geometric features also proves ineffective in handling incorrect correspondences. Finally, FMCW-LIO can generate a neat, complete, and consistent map of the whole tunnel (as shown in Fig. \ref{fig:system_demo}(d)), and the estimated tunnel length is highly close to the real data. Trajectories of algorithms overlaid with a satellite map are shown in Fig. \ref{fig:traj_plot}. It can be found that FMCW-LIO outputs a precious trajectory accurately aligned with the tunnel ends while other algorithms suffer from severe drifts in such structure-degenerated scenes. 

Similarly, STEAM-DICP can only achieve successful running up to the halfway point of each \textit{tunnel} sequence. When the platform executes a U-turn to return, STEAM-DICP fails due to the narrow scene, fast rotations, and a scarcity of keypoints. Despite STEAM-DICP achieving a halfway success in \textit{tunnel} sequences compared to conventional LIO methods, it still exhibits significant drifted errors, especially in the $z$-axis. We only compute the RMSE of the successful segments for STEAM-DICP. An insight from the results is that while the LiDAR-only continuous-time model of STEAM-DICP can partially cope with dynamic motions, it may encounter challenges or fail to handle the narrow workspaces and intense motion on handheld or robot platforms, without the assistance of exteroception-irrelevant IMU sensors.

\begin{table}[!t] 
    \setlength
    \tabcolsep{2.5pt}
    \caption{Average Running Time (Milliseconds) Per LiDAR Scan} 
    \label{tab:table_avg_running_time} 
    \begin{center} 
        \begin{tabular}{cccccc} 
            \toprule 
             & LIO-SAM & DLIO & FAST-LIO2 & STEAM-DICP & FMCW-LIO \\ 
            \midrule
            Structured & 40.08 & 45.93 & 33.53 & 178.25 & 39.71 \\ 
            Degenerated & 28.76 & 36.00 & 20.54 & 85.46 & 21.33 \\ 
            \bottomrule 
        \end{tabular}
    \end{center}
\end{table}

In \textit{tunnel\_02} and \textit{tunnel\_04}, due to high speeds (about 7.0 m/s) of the wheeled platform in the tunnel, the variant (w/o DC) exhibits a significant accuracy decrease which shows the effectiveness of the Doppler compensation in a low-frequency scan pattern. Additionally, in \textit{tunnel\_01} and \textit{tunnel\_03}, the presence of significant vibrations during walking on the handheld platform may lead to a further deterioration of IMU estimation performance when the geometric degeneration occurs, resulting in failures of other algorithms. 
\subsection{Dynamic Point Removal and Static Mapping Results}

\begin{figure}[!t]
       \centering
       \includegraphics[width=0.47\textwidth]{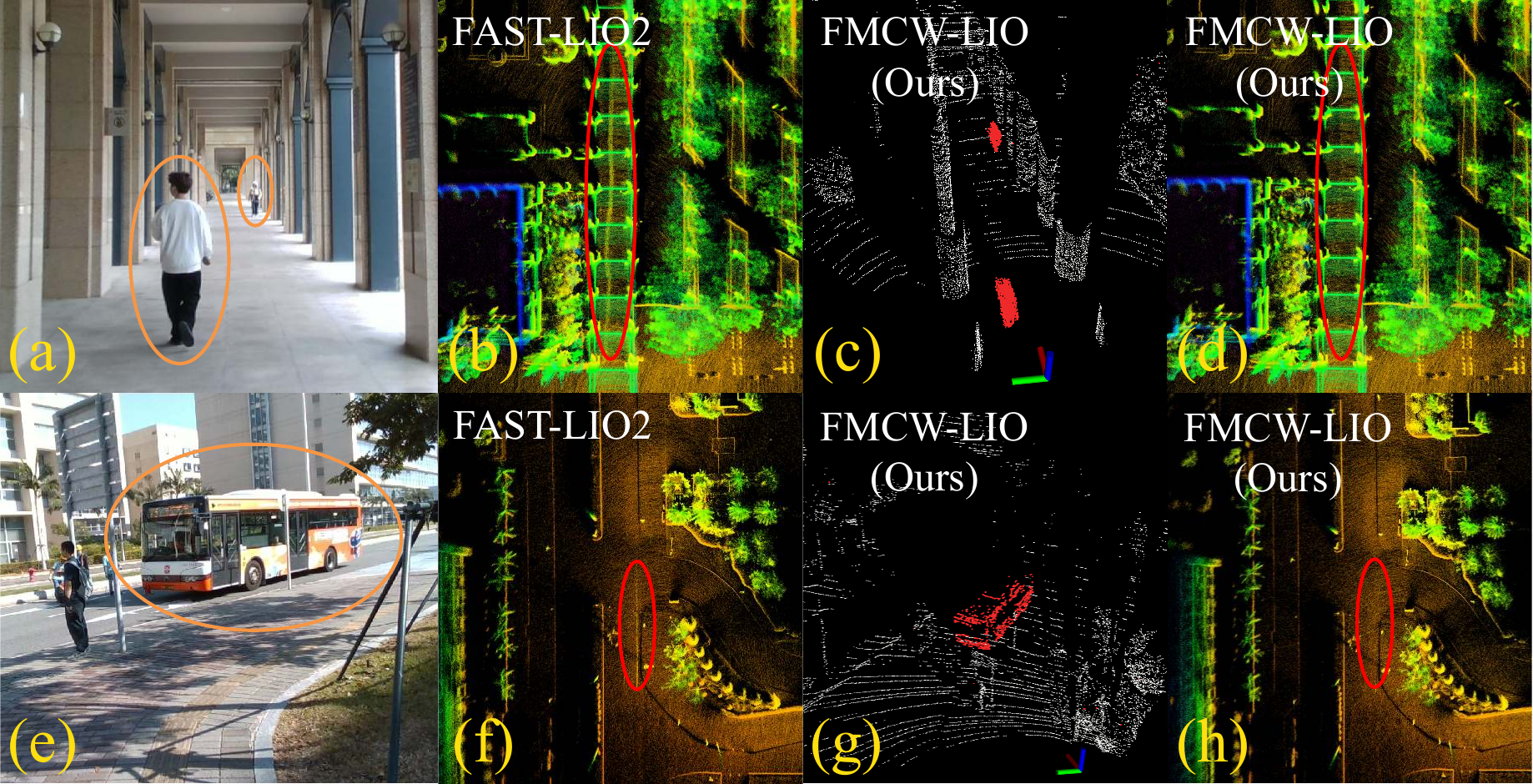}
       \caption{Dynamic point removal and static mapping. (a), (e) Moving persons and bus (orange ellipsoids). (b), (f) Ghost points (red ellipsoids) in the maps of FAST-LIO2. (c), (g) Online dynamic point detection (red points) of FMCW-LIO. (d), (h) Dynamic point removal (no ghost point in red ellipsoids) of FMCW-LIO.} 
       \label{fig:dyn_removal}
    \end{figure}

As described in Section \ref{sec:dyn_removal_static_mapping}, FMCW-LIO is capable of removing dynamic points and generating static maps according to the Doppler residual and the updated LiDAR velocity. The method operates at the point level rather than the semantic or object level, resulting in direct and efficient performance. The results and comparisons of dynamic point detection and static mapping are shown in Fig. \ref{fig:dyn_removal}. It is obvious that, for conventional methods without removal, ghost points and residual blur of moving persons and buses may persist in the online map. Contrarily, FMCW-LIO can detect dynamic points (red points in Fig. \ref{fig:dyn_removal}(c)(g)) in the current scan and finally retain points that are nearly static in the map. As a result, ghost points and blur artifacts do not manifest in the map generated by FMCW-LIO. More online results can be found in the video. 


\section{Conclusion}

In this letter, the Doppler-aided FMCW-LIO is presented for robotic state estimation and mapping. To achieve accurate and robust state estimation, an on-manifold IESEKF is formulated, fusing IMU and FMCW Doppler LiDAR measurements. The motion compensation for Doppler measurements is addressed by the IMU-driven backward propagation. FMCW-LIO utilizes the inherent Doppler-aided LiDAR velocity observation, enabling robust state estimation and static mapping, even in severely narrow or structure-degenerated scenes. Diverse experiments on multiple platforms validate the effectiveness and show the superior performance of FMCW-LIO.


\bibliographystyle{IEEEtran}
\IEEEtriggeratref{21}
\bibliography{ref}

\begin{thebibliography}{10}
\providecommand{\url}[1]{#1}
\csname url@samestyle\endcsname
\providecommand{\newblock}{\relax}
\providecommand{\bibinfo}[2]{#2}
\providecommand{\BIBentrySTDinterwordspacing}{\spaceskip=0pt\relax}
\providecommand{\BIBentryALTinterwordstretchfactor}{4}
\providecommand{\BIBentryALTinterwordspacing}{\spaceskip=\fontdimen2\font plus
\BIBentryALTinterwordstretchfactor\fontdimen3\font minus \fontdimen4\font\relax}
\providecommand{\BIBforeignlanguage}[2]{{%
\expandafter\ifx\csname l@#1\endcsname\relax
\typeout{** WARNING: IEEEtran.bst: No hyphenation pattern has been}%
\typeout{** loaded for the language `#1'. Using the pattern for}%
\typeout{** the default language instead.}%
\else
\language=\csname l@#1\endcsname
\fi
#2}}
\providecommand{\BIBdecl}{\relax}
\BIBdecl

\bibitem{anderson2017ladar}
J.~Anderson, R.~Massaro, J.~Curry, R.~Reibel, J.~Nelson, and J.~Edwards, ``Ladar: frequency-modulated, continuous wave laser detection and ranging,'' \emph{Photogrammetric Engineering \& Remote Sensing}, vol.~83, no.~11, pp. 721--727, 2017.

\bibitem{DICP}
B.~Hexsel, H.~Vhavle, and Y.~Chen, ``{DICP: Doppler Iterative Closest Point Algorithm},'' in \emph{Proceedings of Robotics: Science and Systems}, New York City, NY, USA, June 2022.

\bibitem{LIO-SAM}
T.~Shan, B.~Englot, D.~Meyers, W.~Wang, C.~Ratti, and D.~Rus, ``Lio-sam: Tightly-coupled lidar inertial odometry via smoothing and mapping,'' in \emph{2020 IEEE/RSJ international conference on intelligent robots and systems (IROS)}.\hskip 1em plus 0.5em minus 0.4em\relax IEEE, 2020, pp. 5135--5142.

\bibitem{LIO-Mapping}
H.~Ye, Y.~Chen, and M.~Liu, ``Tightly coupled 3d lidar inertial odometry and mapping,'' in \emph{2019 International Conference on Robotics and Automation (ICRA)}.\hskip 1em plus 0.5em minus 0.4em\relax IEEE, 2019, pp. 3144--3150.

\bibitem{LINS}
C.~Qin, H.~Ye, C.~E. Pranata, J.~Han, S.~Zhang, and M.~Liu, ``Lins: A lidar-inertial state estimator for robust and efficient navigation,'' in \emph{2020 IEEE international conference on robotics and automation (ICRA)}.\hskip 1em plus 0.5em minus 0.4em\relax IEEE, 2020, pp. 8899--8906.

\bibitem{FAST-LIO}
W.~Xu and F.~Zhang, ``Fast-lio: A fast, robust lidar-inertial odometry package by tightly-coupled iterated kalman filter,'' \emph{IEEE Robotics and Automation Letters}, vol.~6, no.~2, pp. 3317--3324, 2021.

\bibitem{FAST-LIO2}
W.~Xu, Y.~Cai, D.~He, J.~Lin, and F.~Zhang, ``Fast-lio2: Fast direct lidar-inertial odometry,'' \emph{IEEE Transactions on Robotics}, vol.~38, no.~4, pp. 2053--2073, 2022.

\bibitem{Faster-LIO}
C.~Bai, T.~Xiao, Y.~Chen, H.~Wang, F.~Zhang, and X.~Gao, ``Faster-lio: Lightweight tightly coupled lidar-inertial odometry using parallel sparse incremental voxels,'' \emph{IEEE Robotics and Automation Letters}, vol.~7, no.~2, pp. 4861--4868, 2022.

\bibitem{Point-LIO}
D.~He, W.~Xu, N.~Chen, F.~Kong, C.~Yuan, and F.~Zhang, ``Point-lio: Robust high-bandwidth light detection and ranging inertial odometry,'' \emph{Advanced Intelligent Systems}, p. 2200459, 2023.

\bibitem{DLIO}
K.~Chen, R.~Nemiroff, and B.~T. Lopez, ``Direct lidar-inertial odometry: Lightweight lio with continuous-time motion correction,'' in \emph{2023 IEEE International Conference on Robotics and Automation (ICRA)}.\hskip 1em plus 0.5em minus 0.4em\relax IEEE, 2023, pp. 3983--3989.

\bibitem{guo2022doppler_vel}
M.~Guo, K.~Zhong, and X.~Wang, ``Doppler velocity-based algorithm for clustering and velocity estimation of moving objects,'' in \emph{2022 7th International Conference on Automation, Control and Robotics Engineering (CACRE)}.\hskip 1em plus 0.5em minus 0.4em\relax IEEE, 2022, pp. 216--222.

\bibitem{CARLA}
A.~Dosovitskiy, G.~Ros, F.~Codevilla, A.~Lopez, and V.~Koltun, ``Carla: An open urban driving simulator,'' in \emph{Conference on robot learning}.\hskip 1em plus 0.5em minus 0.4em\relax PMLR, 2017, pp. 1--16.

\bibitem{gu2022fmcw}
Y.~Gu, H.~Cheng, K.~Wang, D.~Dou, C.~Xu, and H.~Kong, ``Learning moving-object tracking with fmcw lidar,'' in \emph{2022 IEEE/RSJ International Conference on Intelligent Robots and Systems (IROS)}.\hskip 1em plus 0.5em minus 0.4em\relax IEEE, 2022, pp. 3747--3753.

\bibitem{Picking_up_Speed}
Y.~Wu, D.~J. Yoon, K.~Burnett, S.~Kammel, Y.~Chen, H.~Vhavle, and T.~D. Barfoot, ``Picking up speed: Continuous-time lidar-only odometry using doppler velocity measurements,'' \emph{IEEE Robotics and Automation Letters}, vol.~8, no.~1, pp. 264--271, 2022.

\bibitem{Need_for_Speed}
D.~J. Yoon, K.~Burnett, J.~Laconte, Y.~Chen, H.~Vhavle, S.~Kammel, J.~Reuther, and T.~D. Barfoot, ``Need for speed: Fast correspondence-free lidar-inertial odometry using doppler velocity,'' in \emph{2023 IEEE/RSJ International Conference on Intelligent Robots and Systems (IROS)}.\hskip 1em plus 0.5em minus 0.4em\relax IEEE, 2023, pp. 5304--5310.

\bibitem{yoneda2023EOT}
M.~Yoneda, K.-M. Dahl{\'e}n, and T.~Ogawa, ``Extended object tracking with doppler velocity-based point registration,'' in \emph{2023 IEEE Symposium Sensor Data Fusion and International Conference on Multisensor Fusion and Integration (SDF-MFI)}.\hskip 1em plus 0.5em minus 0.4em\relax IEEE, 2023, pp. 1--8.

\bibitem{HeLiPR}
M.~Jung, W.~Yang, D.~Lee, H.~Gil, G.~Kim, and A.~Kim, ``Helipr: Heterogeneous lidar dataset for inter-lidar place recognition under spatial and temporal variations,'' 2023.

\bibitem{rio}
C.~Doer and G.~F. Trommer, ``An ekf based approach to radar inertial odometry,'' in \emph{2020 IEEE International Conference on Multisensor Fusion and Integration for Intelligent Systems (MFI)}.\hskip 1em plus 0.5em minus 0.4em\relax IEEE, 2020, pp. 152--159.

\bibitem{michalczyk2022tightly-coupled_rio}
J.~Michalczyk, R.~Jung, and S.~Weiss, ``Tightly-coupled ekf-based radar-inertial odometry,'' in \emph{2022 IEEE/RSJ International Conference on Intelligent Robots and Systems (IROS)}.\hskip 1em plus 0.5em minus 0.4em\relax IEEE, 2022, pp. 12\,336--12\,343.

\bibitem{Milli-RIO}
Y.~Almalioglu, M.~Turan, C.~X. Lu, N.~Trigoni, and A.~Markham, ``Milli-rio: Ego-motion estimation with low-cost millimetre-wave radar,'' \emph{IEEE Sensors Journal}, vol.~21, no.~3, pp. 3314--3323, 2020.

\bibitem{ng2021continuous}
Y.~Z. Ng, B.~Choi, R.~Tan, and L.~Heng, ``Continuous-time radar-inertial odometry for automotive radars,'' in \emph{2021 IEEE/RSJ International Conference on Intelligent Robots and Systems (IROS)}.\hskip 1em plus 0.5em minus 0.4em\relax IEEE, 2021, pp. 323--330.

\bibitem{4D_iRIOM}
Y.~Zhuang, B.~Wang, J.~Huai, and M.~Li, ``4d iriom: 4d imaging radar inertial odometry and mapping,'' \emph{IEEE Robotics and Automation Letters}, 2023.

\bibitem{IKFoM-arXiv}
D.~He, W.~Xu, and F.~Zhang, ``Kalman filters on differentiable manifolds,'' \emph{arXiv preprint arXiv:2102.03804}, 2021.

\bibitem{IMU_Allan}
N.~El-Sheimy, H.~Hou, and X.~Niu, ``Analysis and modeling of inertial sensors using allan variance,'' \emph{IEEE Transactions on instrumentation and measurement}, vol.~57, no.~1, pp. 140--149, 2007.

\bibitem{LOAM}
J.~Zhang and S.~Singh, ``Loam: Lidar odometry and mapping in real-time.'' in \emph{Robotics: Science and systems}, vol.~2, no.~9.\hskip 1em plus 0.5em minus 0.4em\relax Berkeley, CA, 2014, pp. 1--9.

\bibitem{aeva2023}
\BIBentryALTinterwordspacing
{Aeva Inc. Aeries II}, Accessed Aug. 8, 2023. [Online]. Available: \url{{https://www.aeva.com/aeries-ii/}}
\BIBentrySTDinterwordspacing

\bibitem{wang2020lidar_iris}
Y.~Wang, Z.~Sun, C.-Z. Xu, S.~E. Sarma, J.~Yang, and H.~Kong, ``Lidar iris for loop-closure detection,'' in \emph{2020 IEEE/RSJ International Conference on Intelligent Robots and Systems (IROS)}.\hskip 1em plus 0.5em minus 0.4em\relax IEEE, 2020, pp. 5769--5775.

\bibitem{dellaert2012GTSAM}
F.~Dellaert, ``Factor graphs and gtsam: A hands-on introduction,'' \emph{Georgia Institute of Technology, Tech. Rep}, vol.~2, p.~4, 2012.

\end{thebibliography}


\newpage

\section*{\textbf{Supplementary Framework Documentation}}

    \begin{figure*}[!b]
       \centering
       \includegraphics[width=0.99\textwidth]{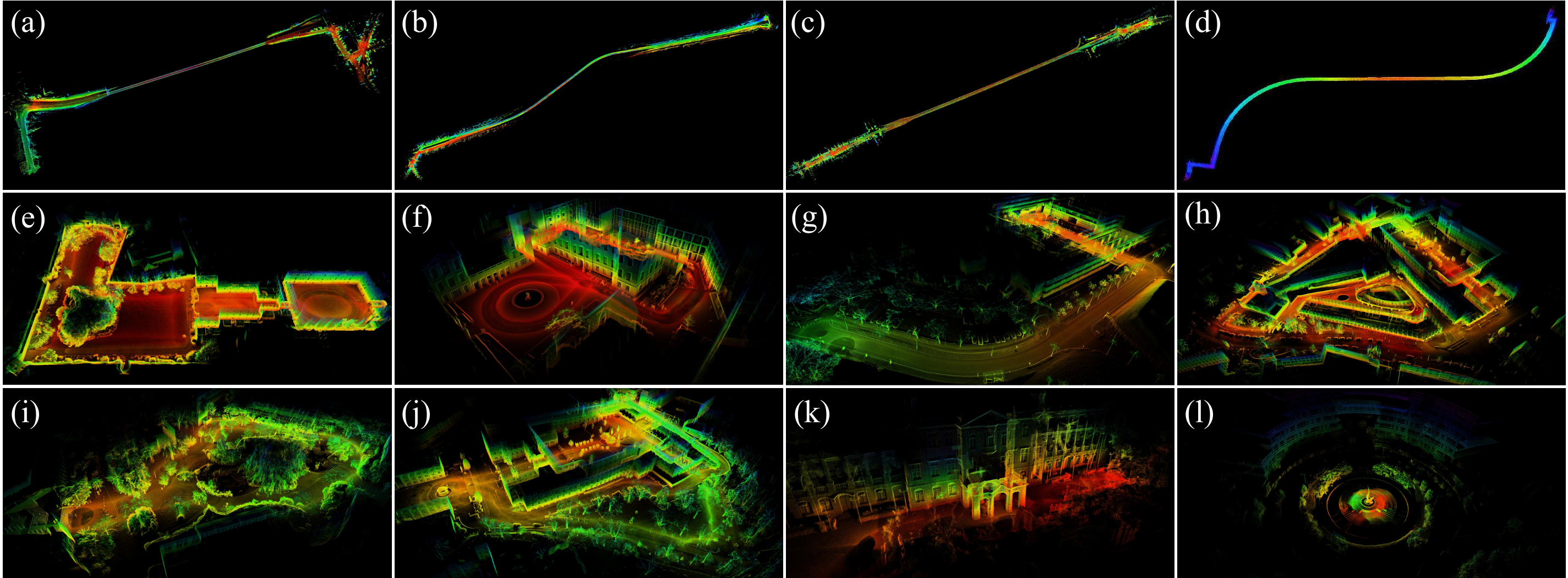}
       \caption{Representative mapping results of {\href{https://doi.org/10.1109/LRA.2024.3396636}{\FMCWLIO}} on different typical datasets. (a), (b) Boreas-RT Dataset. (c) HeRCULES Dataset. (e), (f) Newer College Dataset. (d) FMCW-LIO Dataset. (g)--(j) MCD. (k) FAST-LIO2 Dataset. (l) FAST-LIVO2 Dataset.}
       \label{fig:mapping_results}
    \end{figure*}

{\href{https://doi.org/10.1109/LRA.2024.3396636}{\FMCWLIO}} is a unified and extensible LiDAR-inertial odometry framework that supports both 4D Doppler LiDARs and conventional 3D LiDARs. Built on a modular architecture design, the key framework components can be configured, extended, or replaced independently. Beyond the Doppler-aided estimation, {\href{https://doi.org/10.1109/LRA.2024.3396636}{\FMCWLIO}} also offers the following features:
    \begin{itemize}
        \item \textbf{Doppler Velocity Compensation}: A Doppler compensation scheme aligns Doppler velocity measurements into a unified spatiotemporal frame. Beyond motion compensation, Doppler compensation essentially reveals the interrelation between the Doppler velocities of a static point with respect to arbitrary different frames.
    
        \item \textbf{Flexible Initialization}: Multiple initialization algorithm and strategies are supported, including conventional stationary initialization and {\href{https://doi.org/10.1109/LRA.2024.3490395}{\FreeInit}}, which delivers accurate initial states even in the presence of aggressive motion, moving objects, and structure degeneracy.
    
        \item \textbf{Multiple Integration and Multiple Motion Compensation Schemes}: Various integration and motion compensation methods are provided, including second-order Runge-Kutta method (RK2), third-order Runge-Kutta method (RK3), fourth-order Runge-Kutta method (RK4), mechanization, and ACI2.
    
        \item \textbf{Robust Velocity Estimation}: Multiple algorithms are integrated to improve the resilience and accuracy of velocity estimation, including least squares with RANSAC, covariance, and robust kernels.
    
        \item \textbf{Multiple Map Structures}: Different map structures are supported, including {\href{https://doi.org/10.1109/TRO.2022.3141876}{ikd-Tree}} and {\href{https://doi.org/10.1109/LRA.2022.3152830}{iVox}}, enabling flexible and efficient mapping under varying conditions.
    
        \item \textbf{Multiple LiDAR Types}: The framework is compatible with both 4D Doppler LiDARs and conventional 3D LiDARs, facilitating evaluation and deployment across different LiDAR sensing configurations.
    \end{itemize}

{\href{https://doi.org/10.1109/LRA.2024.3396636}{\FMCWLIO}} is tested on numerous typical LiDAR-inertial datasets covering different LiDAR sensors, platform types, motion patterns, and environmental conditions, including:
    \begin{enumerate}
        \item {\href{https://www.boreas.utias.utoronto.ca/\#/boreasRT/}{Boreas-RT Dataset}}{\footnote{\href{https://www.boreas.utias.utoronto.ca/\#/boreasRT/}{https://www.boreas.utias.utoronto.ca/\#/boreasRT/}}};
        
        \item {\href{https://sites.google.com/view/herculesdataset/}{HeRCULES Dataset}}{\footnote{\href{https://sites.google.com/view/herculesdataset/}{https://sites.google.com/view/herculesdataset/}}};
        
        \item {\href{https://ori-drs.github.io/newer-college-dataset/}{Newer College Dataset}}{\footnote{\href{https://ori-drs.github.io/newer-college-dataset/}{https://ori-drs.github.io/newer-college-dataset/}}};
        
        \item {\href{https://dynamic.robots.ox.ac.uk/datasets/oxford-spires/}{Oxford Spires Dataset}}{\footnote{\href{https://dynamic.robots.ox.ac.uk/datasets/oxford-spires/}{https://dynamic.robots.ox.ac.uk/datasets/oxford-spires/}}};
        
        \item {\href{https://mcdviral.github.io/}{Multi-Campus Dataset (MCD)}}{\footnote{\href{https://mcdviral.github.io/}{https://mcdviral.github.io/}}};
        
        \item {\href{https://ntu-aris.github.io/ntu_viral_dataset/}{NTU VIRAL Dataset}}{\footnote{\href{https://ntu-aris.github.io/ntu_viral_dataset/}{https://ntu-aris.github.io/ntu\_viral\_dataset/}}};
        
        \item {\href{https://hilti-challenge.com/dataset-2022.html}{Hilti SLAM Challenge Dataset 2022}}{\footnote{\href{https://hilti-challenge.com/dataset-2022.html}{https://hilti-challenge.com/dataset-2022.html}}};
        
        \item {\href{https://github.com/IPNL-POLYU/UrbanNavDataset}{UrbanNav Dataset}}{\footnote{\href{https://github.com/IPNL-POLYU/UrbanNavDataset}{https://github.com/IPNL-POLYU/UrbanNavDataset}}};
        
        \item {\href{https://github.com/hku-mars/FAST_LIO}{FAST-LIO2 Dataset}}{\footnote{\href{https://github.com/hku-mars/FAST_LIO}{https://github.com/hku-mars/FAST\_LIO}}};
        
        \item {\href{https://github.com/hku-mars/FAST-LIVO2}{FAST-LIVO2 Dataset}}{\footnote{\href{https://github.com/hku-mars/FAST-LIVO2}{https://github.com/hku-mars/FAST-LIVO2}}}.
    \end{enumerate}
Extensive test experiments are performed to validate the framework across heterogeneous sensing configurations and environmental conditions. {\href{https://doi.org/10.1109/LRA.2024.3396636}{\FMCWLIO}} demonstrates consistently robust and accurate odometry and mapping results with extremely efficient performance, within both well-structured and structure-degenerated scenes. Representative mapping results on selected sequences are presented in Fig. \ref{fig:mapping_results}.

In addition, the typical 4D FMCW Doppler LiDAR-inertial sequences collected in this work are also released as the \href{https://github.com/IMRL/FMCW-LIO}{FMCW-LIO Dataset} and the \href{https://github.com/IMRL/Free-Init}{Free-Init Dataset}. FMCW-LIO Dataset provides five handheld sequences across campus, street, and tunnel scenes, covering both well-structured and structure-degenerated environments. Free-Init Dataset provides four handheld and vehicular sequences recorded under dynamic initial motions for the qualitative evaluation of dynamic initialization. Characteristics and download links of the sequences are listed and detailed on the respective project pages.

\end{document}